\documentclass[10pt,twocolumn,letterpaper]{article}

 \usepackage[pagenumbers]{cvpr} 
\usepackage{algorithm}
\usepackage{algpseudocode}
\usepackage{graphicx}
\usepackage{array} 

\definecolor{cvprblue}{rgb}{0.21,0.49,0.74}
\usepackage[pagebackref,breaklinks,colorlinks,allcolors=cvprblue]{hyperref}

\def\paperID{11122} 
\def\confName{CVPR}
\def\confYear{2025}

\title{ColorEdit: Training-free Image-Guided Color editing with diffusion model}

\author{
    Xingxi Yin \quad Zhi Li \quad Jingfeng Zhang \quad Chenglin Li \quad Yin Zhang\\
    Zhejiang University\\
    Hangzhou, China\\
    {\tt\small \{12321103\}@zju.edu.cn}
}

\begin{document}
\maketitle

\begin{abstract}
Text-to-image (T2I) diffusion models, with their impressive generative capabilities, have been adopted for image editing tasks, demonstrating remarkable efficacy.  However, due to attention leakage and collision between the cross-attention map of the object and the new color attribute from the text prompt, text-guided image editing methods may fail to change the color of an object, resulting in a misalignment between the resulting image and the text prompt.  In this paper, we conduct an in-depth analysis on the process of text-guided image synthesizing and what semantic information different cross-attention blocks have learned. We observe that the visual representation of an object is determined in the up-block of the diffusion model in the early stage of the denoising process, and color adjustment can be achieved through value matrices alignment in the cross-attention layer. Based on our findings, we propose a straightforward, yet stable, and effective image-guided method to modify the color of an object without requiring any additional fine-tuning or training. Lastly, we present a benchmark dataset called COLORBENCH, the first benchmark to evaluate the performance of color change methods. Extensive experiments validate the effectiveness of our method in object-level color editing and surpass the performance of popular text-guided image editing approaches in both synthesized and real images.
\end{abstract}    
\section{Introduction}
\label{sec:intro}
\begin{figure*}[ht]
    \centering
    \includegraphics[width=1\textwidth]{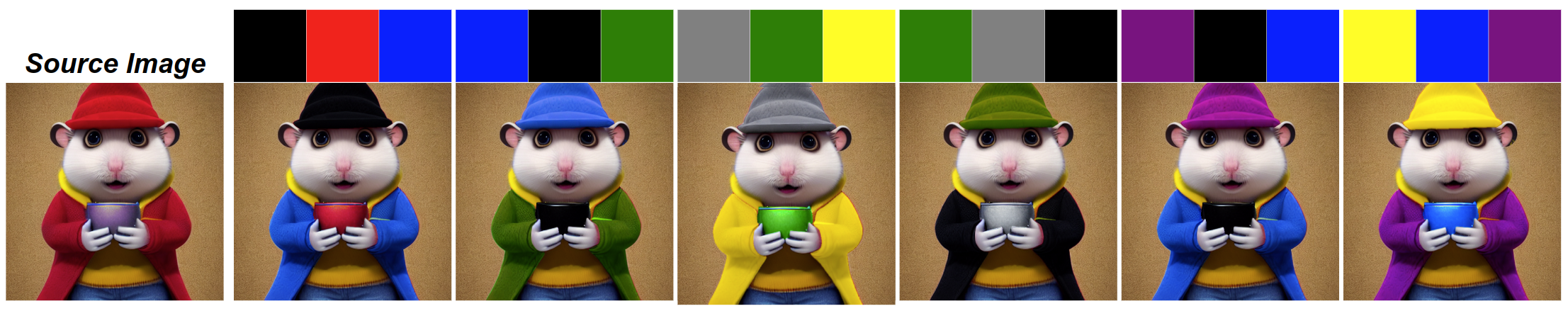}
    \caption{ \textbf{Multi-Object color editing.} \textnormal{Each outcome image is changing the color of the hat first and then changing the color of the bowl and coat, using the associated reference color image. } }
    \label{fig:teaser}
\end{figure*}

\begin{figure}[ht]
    \centering
    \includegraphics[width=0.5\textwidth]{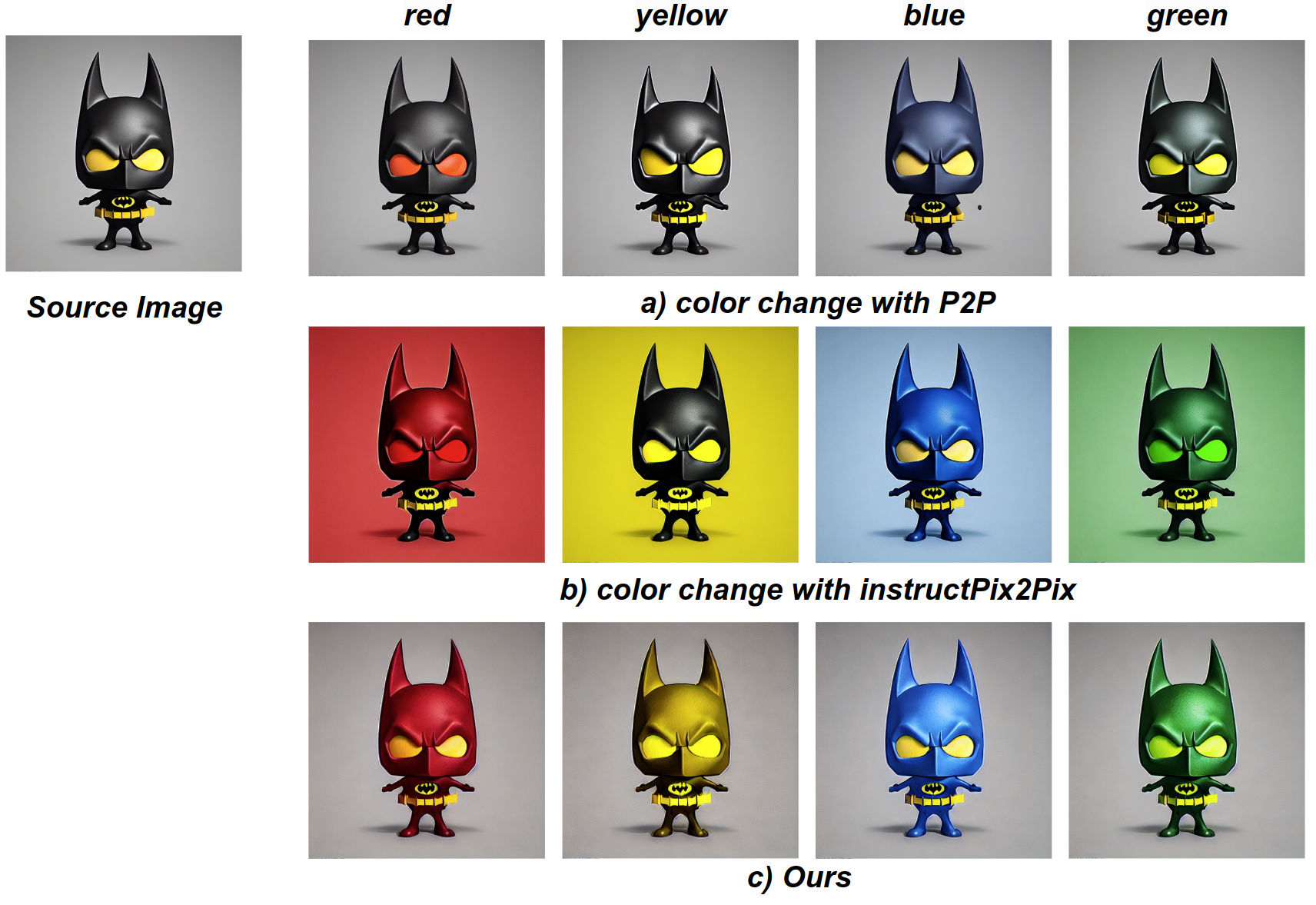}
    \caption{ \textbf{Example of color change.} \textnormal{Text-guided editing methods may fail to change the color of an object while maintaining the structure of it or the background information. } }
    \label{fig:text_guided_color_change}
\end{figure}

Color is one of the most important visual perceptions for humans. The color of an object significantly influences the emotional responses and perceptions of a person toward it, which makes color a key element in both functional and aesthetic design decisions across industries, particularly in the design field. With the development of diffusion models, some studies \cite{huang2022unicolor, weng2024cad, liang2024control} have applied the stable diffusion (SD) model to the colorization task, which involves adding color to grayscale or black-and-white images. However, currently, no studies specifically investigate the task of color change. Although some methods \cite{hertz2022prompt, tumanyan2023plug, liu2024towards, brooks2023instructpix2pix, geng2024instructdiffusion, fu2023guiding} demonstrate the capability to modify an object's color, those text-guided techniques often fail to change the color of an object as expected, as shown in Fig.~\ref{fig:text_guided_color_change}. 

In this paper, we conduct an in-depth exploration on the cross-attention layer, which aligns and transforms text information into synthesized images. Specifically, we visualize the cross-attention maps of objects in different blocks of the model through different stages of the denoising process to elucidate how textual information directs the generation of images. We identify the semantic information captured by various cross-attention blocks and demonstrate when and where an object's shape, contour, and texture are established. For the unsuccessful outcome of text-guided color editing, we argue that there are two main factors: (a) the imprecise distribution of color attribute attention weights on the spatial area, called cross-attention leakage. (b) The collision of information on attributes in the cross-attention map from the original object and the color term in the target prompt. We find that, compared to altering the Key matrices in the cross-attention layer, modifying the Value matrices of the target image results in a more stable color change effect. Based on our findings, we introduce a simplified yet stable and effective method called training-free Image-Guided Color Editing. Our method performs object color editing by aligning the Value matrices of the target image with the Value matrices extracted from a reference color image in specific cross-attention layers of the diffusion model in the early stages of the denoising process (see an example in Fig.~\ref{fig:teaser}). 

\section{Related work}
\label{sec:related_work}

\begin{figure*}[ht]
    \centering
  \includegraphics[width=\textwidth]{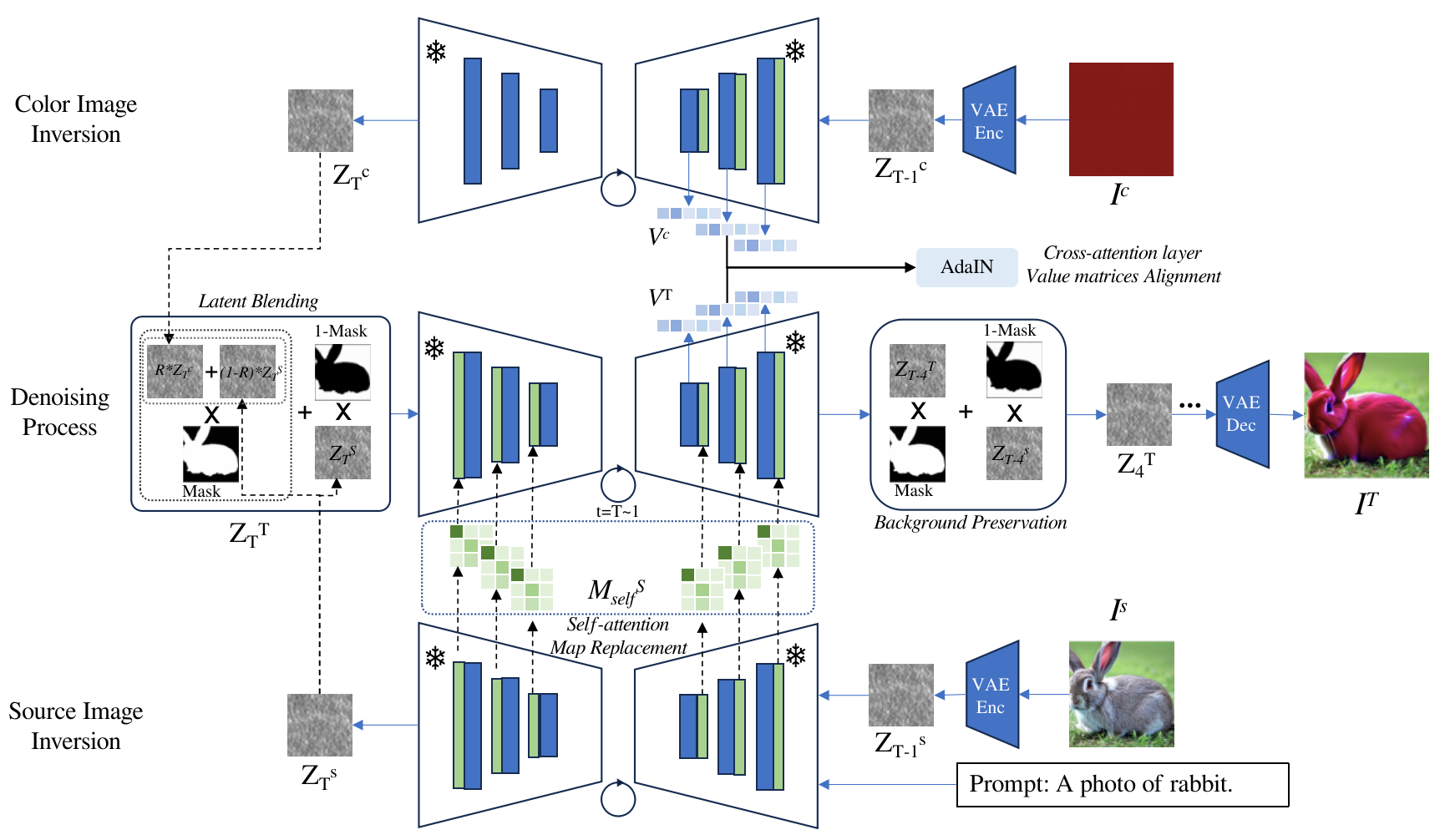}
  \caption{ \textbf{The Image-guided Color Editing Framework. }\textnormal{Our framework including: (a) Color Image Inversion. The reference color image is inverted to initial noise and extracted the Value matrices of the cross-attention layers. (b) Source Image inversion. Extracting the self-attention map and latent $Z^s_T, ..., Z^s_0$ of the source image.  (c) Denoising Process, which including latent blending at the begining,  cross-attention layer value matrices alignment in the early stage, self-attention map replacement through the whole process, and background preserve in the last few steps. }}
   \label{fig: Framework_pipeline}
\end{figure*}

Significant advances in diffusion-based text-guided image generation methods \cite{ho2022classifier, nichol2021glide, saharia2022photorealistic,rombach2022high, ramesh2022hierarchical, meng2023distillation, gu2022vector, podell2023sdxl} have led to the rapid adoption of these methods for various generative visual tasks, including image editing, which includes semantic editing, structural modifying, and stylistic editing.  Semantic editing involves altering the fundamental meaning or conceptual representation of an image. This may include tasks such as adding \cite{xie2023smartbrush, kawar2023imagic, wang2023imagen, lu2023tf, chen2023anydoor, brooks2023instructpix2pix, geng2024instructdiffusion, parmar2023zero, avrahami2023blended, goel2023pair, fu2023guiding} , removing \cite{yildirim2023inst, geng2024instructdiffusion, fu2023guiding}, or replacing objects  \cite{xie2023smartbrush, kawar2023imagic, hertz2022prompt, couairon2022diffedit,  liu2024towards, yu2023inpaint, chen2023anydoor, brooks2023instructpix2pix, geng2024instructdiffusion, fu2023guiding, parmar2023zero, avrahami2023blended, kim2022diffusionclip, goel2023pair, mokady2023null} , and changing the background \cite{li2023layerdiffusion, avrahami2023blended}. Structural editing modifies the composition of an image, including repositioning \cite{epstein2023diffusion, chen2023anydoor}, changing the size and shape \cite{patashnik2023localizing, epstein2023diffusion, li2023layerdiffusion, goel2023pair} , altering actions and poses \cite{li2023layerdiffusion, kawar2023imagic, cao2023masactrl, huang2023kv}, and adjusting the perspective or viewpoint \cite{cao2023masactrl} of an object. Stylistic editing involves modifying the artistic style or aesthetic presentation of an image, including color change \cite{hertz2022prompt,  liu2024towards, tumanyan2023plug,  brooks2023instructpix2pix, fu2023guiding, geng2024instructdiffusion, kawar2023imagic, mokady2023null}, texture change \cite{tumanyan2023plug}, and style change \cite{parmar2023zero, kawar2023imagic, sohn2023styledrop, geng2024instructdiffusion, tumanyan2023plug, zhao2023null, kim2022diffusionclip, wang2023stylediffusion, xu2024cyclenet, mokady2023null}.

For color change task, according to their learning strategies, text-guided image editing methods can be classified into training-based approaches \cite{brooks2023instructpix2pix, geng2024instructdiffusion, fu2023guiding}, testing-time fine-tuning approaches \cite{kawar2023imagic, mokady2023null}, and training- and fine-tuning-free approaches \cite{hertz2022prompt,  tumanyan2023plug, cao2023masactrl, liu2024towards}.  Training on a constructed image-text-image dataset, InstructPix2Pix \cite{brooks2023instructpix2pix}, InstructDiffusion \cite{geng2024instructdiffusion}, and MGIE \cite{fu2023guiding} train a new diffusion model, allowing users to edit images using natural language instructions. Imagic \cite{kawar2023imagic} approximates the input image's text embedding through tuning, then edits the image by interpolating between this embedding and the target text embedding.  Null text inversion \cite{mokady2023null} employs an optimization method to reconstruct the image and utilizes P2P \cite{hertz2022prompt} for image editing. P2P \cite{hertz2022prompt} observes that cross-attention layers capture interactions between pixel structures and prompt words, enabling semantic image editing through cross-attention map manipulation.  PnP \cite{tumanyan2023plug} demonstrated that the structure of the image can be preserved by manipulating spatial features and self-attention maps. FreePromptEditing \cite{liu2024towards} explores cross-attention and self-attention mechanisms, showing that cross-attention maps often carry object attribution information, leading to editing failures, while self-attention maps are essential for preserving geometric and shape details during transformations. Although all of these methods are capable of editing the color of an object in an image, none are specifically designed to address the task of color change. In this paper, we introduce a training-free method which utilizes a reference color image to modify the color of an object. Unlike training-based or testing-time fine-tuning approaches \cite{brooks2023instructpix2pix, geng2024instructdiffusion, kawar2023imagic, mokady2023null},  our method operates without requiring any fine-tuning process. The most related work to ours is training and fine-tuning free approaches P2P \cite{hertz2022prompt}, PnP \cite{tumanyan2023plug}, and FreePromptEditing \cite{liu2024towards}. However, due to attention leakage and attribution collision between the cross-attention map of the original object and the new attribute in the text prompt, those methods may fail on the color change task. 

\section{Preliminaries}
\label{preliminaries}

\textbf{3.1 Latent Diffusion model}

Diffusion models \cite{ho2020denoising, song2020score, sohl2015deep} involve two Markov chain processes. The forward diffusion process is designed to transform a dataset's distribution into a specified distribution, such as the Gaussian distribution.  The denoising process removes noise by sampling the data from a learned Gaussian distribution. The goal of the denoising process is to learn to reverse the forward diffusion process, ultimately reconstructing a distribution that closely mirrors the original.  Our approach takes advantage of the advanced text-conditioned model, Stable Diffusion (SD) \cite{rombach2022high}. The denoising backbone is structured as a time-conditional U-Net \cite{ronneberger2015u}. 

\textbf{3.2 Attention Mechanism in LDM}

The U-Net in the Stable Diffusion model is an Encoder-Decoder architecture comprising an encoder, a decoder, and a connection module that integrates these components. The encoder has three CrossAttnDownBlocks, while the decoder utilizes three CrossAttnUpBlocks, and the connection module is a CrossAttnMidBlock. Each Cross-Attention block in the U-Net consists of a series of basic blocks. Each basic block includes a residual block, a self-attention module, and a cross-attention module. The core of the self-attention module and the cross-attention module lies in the attention mechanism \cite{ashish2017attention}, which can be formulated as follows:
\[Attention(Q, K, V) =softmax\left(\frac{QK^{T}}{\sqrt{d_k}}\right)V, \quad (1) \]
Here, $Q$ represents the Query matrices projected from the spatial features, while $K$ and $V$ denote the Key and Value matrices, which are projected from either the spatial features (in the self-attention module) or the textual embeddings (in the cross-attention module). $d_k$ is the dimension of $K$. 
\section{Method}
\label{sec:method}

Given a source image $I^{s}$ or a text prompt $P$, an interest object $O$, and a reference color image $I^c$,  our objective is to change the color of the object and output a target image $I^{T}$.  In the target image, the object's color should align with the reference color image, while the shape, contour, and achromatic texture of the object should be the same in the source image.  Moreover, the background areas of the image should remain unchanged. To solve this challenge, we propose a training-free framework that extracts Value matrices $V^c$ from the cross-attention layer of a reference color image and uses them to align the corresponding Value $V^T$ of the target image during the denoising process to modify the color of the object.  The framework pipeline is given in Fig.~\ref{fig: Framework_pipeline} and the pseudocode algorithms to edit the color of an object on generated and real images are given in Section 7.1 of the Supplementary Material.  

\textbf{4.1 Cross-Attention Mechanism in LDM}

\begin{figure}
    \centering
    \includegraphics[width=0.5\textwidth]{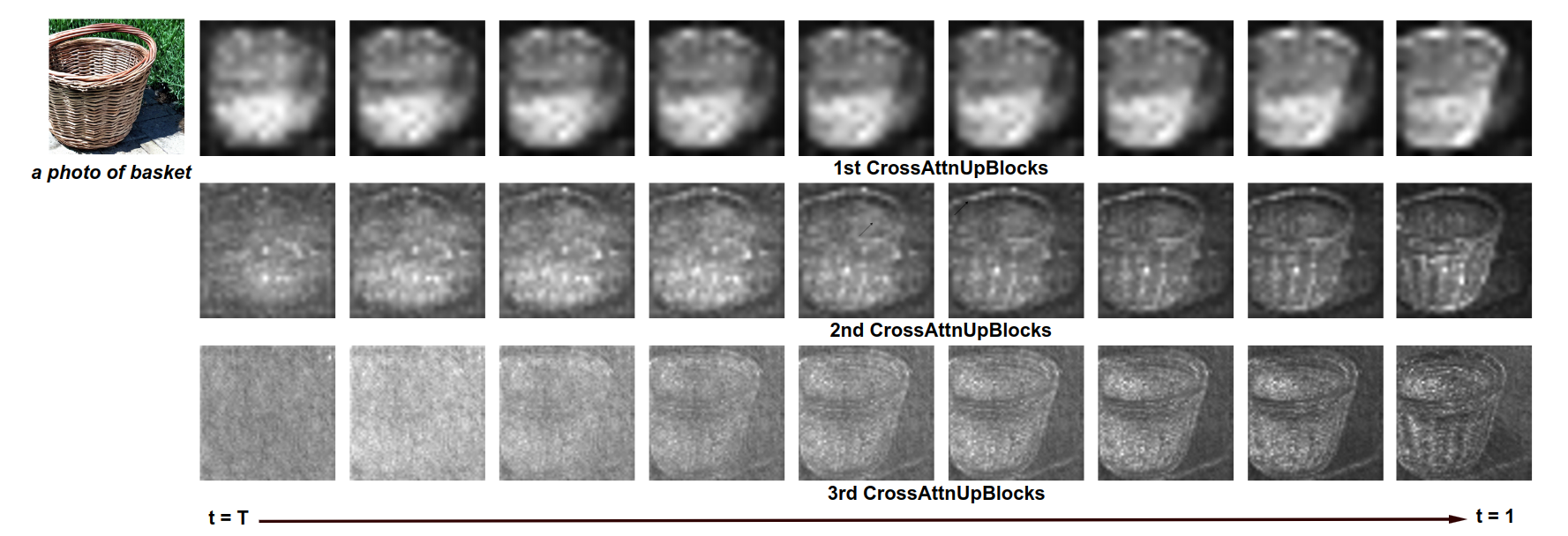}
    \caption{\textbf{Cross-attention map of object in different denoising steps in the decoder of U-net. } We visualize the cross-attention maps of an object at various diffusion steps within the U-Net decoder. } 
    \label{fig:cross-attention-map-object-different-timestep-decoder-unet}
\end{figure}

In text-guided image generation, the text guiding process utilizes the mechanism of cross-attention \cite{vaswani2017attention} within a low-dimensional latent space. This mechanism aligns the image pixel with the semantic meaning of the prompt words. Let the text prompt be $P$, the textual embedding be $\phi(P)$, the noisy vector in the latent space at the diffusion step $t$ be $z_t$, the deep spatial features of the noisy vector be $\phi(z_t)$. The Query $Q$, Key $K$, and Value $V$ in the cross-attention layer are obtained through learned linear projections function $f_{Q}$,  $f_{K}$ and  $f_{V}$ separately.
\[
Q = {f_{Q}}(\phi(z_t)),  K = {f_{K}} (\phi(P)), V = {f_{V}} (\phi(P)), \quad (2) 
\]
It is obvious that textual information encoding is fed into the linear projection function $f_{K}$, $f_{V}$, which means that the text prompt affected the image synthesis process directly through the Key and Value matrices, while Q is derived from the previous layer. The query $Q$ and the key $K$ are then used to calculate the attention map, which determines the spatial layout and geometry of the generated image. In this paper, we investigate the interaction between pixel spatial structures and subjects in the prompts in different blocks of the diffusion model.  As can be seen in Fig.~\ref{fig:cross-attention-map-of-object-different-block}, the shape, contour, and texture of an object in the generated image are determined in the U-Net decoder. Specifically, the shape, contour, and texture of the object are defined in the 1st, 2nd, and 3rd CrossAttnUpBlocks. More detailed results of the cross-attention map of the object within different cross-attention blocks can be seen in Section 7.2. In Fig.~\ref{fig:cross-attention-map-object-different-timestep-decoder-unet}, we can observe that the shape is established first, followed by the contour and texture, which are refined until the end of the process. The main shape and contour of an object are established in the early stage of the denoising process. 

\begin{figure}
    \centering
    \includegraphics[width=0.5\textwidth]{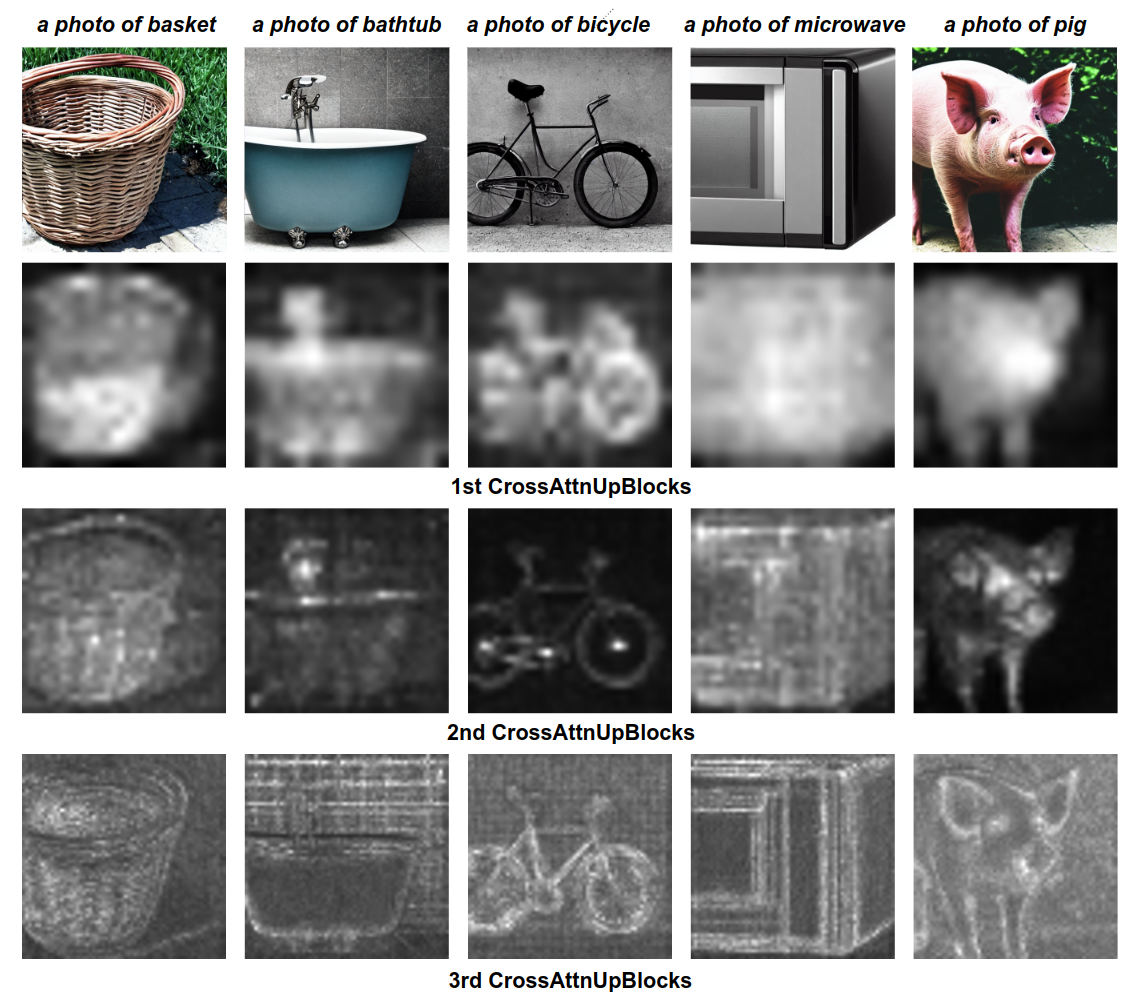}
    \caption{\textbf{Cross-attention map of objects in the decoder of U-net.} We visualize the average cross-attention maps of various objects across all timesteps. As observed, the shape, contour, and texture of an object are determined in the U-Net decoder. }    \label{fig:cross-attention-map-of-object-different-block}
\end{figure}

\begin{figure}
    \centering
    \includegraphics[width=0.5\textwidth]{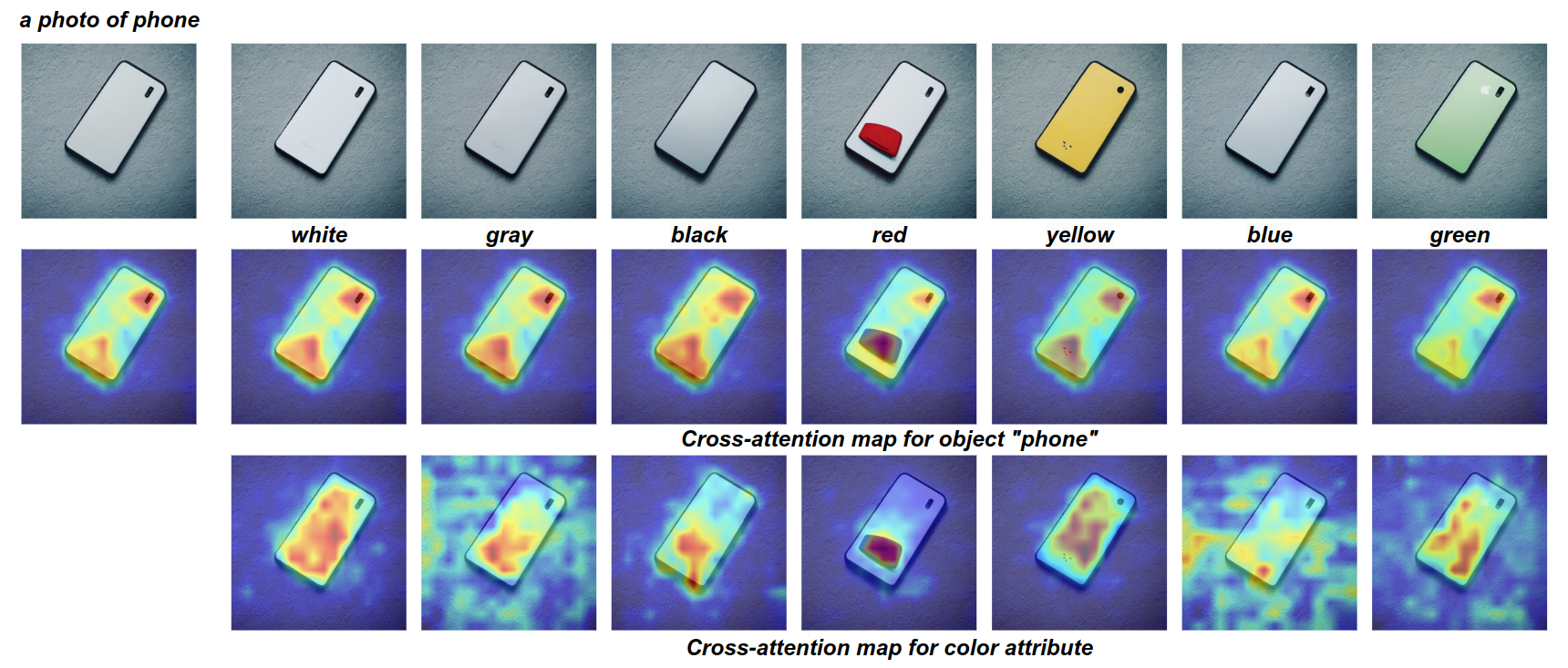}

    \caption{\textbf{Cross-attention maps for object and color attributes using P2P to edit the color. } As indicated, the mis-distribution or collision of the attention map between the color attribute and the object are the reasons fail to change the color of an object.} 
    \label{fig:p2p_cross_attention_map_object_color}
\end{figure}

\begin{figure}
    \centering
    \includegraphics[width=0.5\textwidth]{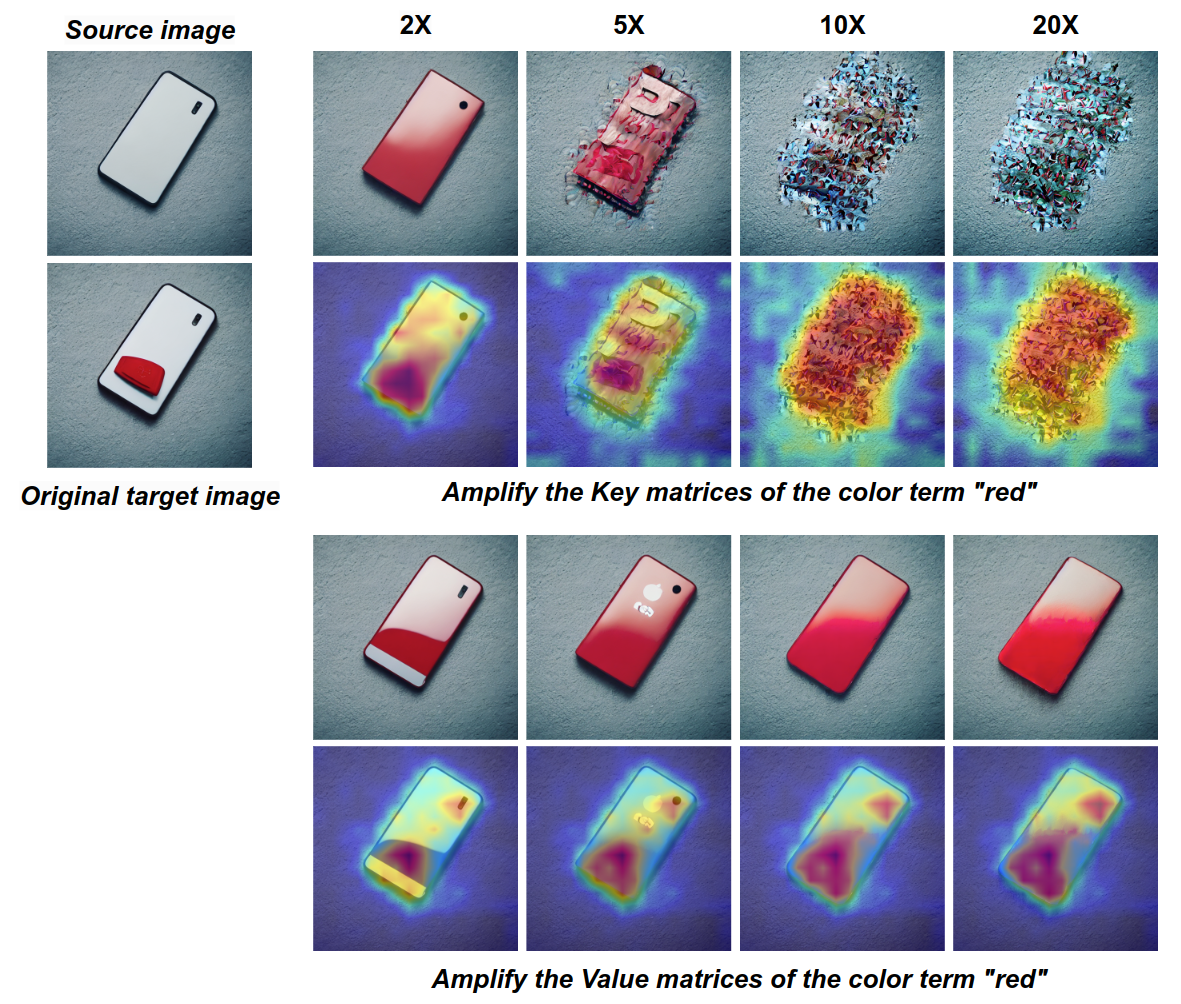}
    \caption{\textbf{Resulting image and cross-attention after amplifying the Key matrices and Value matrices of color term when using P2P to change the color of the phone to red.}  where "*x" means * time of the original value. As illustrated, compared to altering the Key, modifying the Value of the target image can lead to more stable color alterations.} 
    \label{fig:Output_image_and_Cross_attention_map_of_color_attribute}
\end{figure}

When changing color using text-guided training-free image editing methods \cite{hertz2022prompt,  liu2024towards}, the color editing task is performed by adding a color term to describe the object. During the denoising process, the cross-attention map of this color is injected into the corresponding part of the cross-attention map from the source prompt. However, as shown in Fig.~\ref{fig:p2p_cross_attention_map_object_color}, those methods may fail due to the misdistribution of the attention map between the color attribute and the object (e.g. when trying to change the phone color to red, but it only changes in the most active areas of the color attribute cross-attention map). Additionally, collision of attribute information can occur between the cross-attention maps of the object and the color (e.g., when trying to change the phone color to white and black, where the color attribute’s cross-attention map is well distributed, but the color change is ineffective), as the cross-attention map of the object retains its original information of the color attribute \cite{liu2024towards}. For training-based text-guided image editing methods \cite{brooks2023instructpix2pix, geng2024instructdiffusion, fu2023guiding}, the color change task is performed directly through text prompts, and attention leakage also exists (see an example in Section 7.3). Successful color change requires addressing cross-attention leakage between the color attribute and object in spatial space, as well as attribute collisions in the cross-attention map. A straightforward method is to amplify the Key or Value of the color descriptor in the target prompt to increase the color information.  As illustrated in Fig.~\ref{fig:Output_image_and_Cross_attention_map_of_color_attribute}, amplifying the Key matrices of the color red will modify the phone's structure. In contrast, adjusting the Value matrices leads to a more stable color change, but it still changes the structure of the phone to some extent. However, this method doesn't work for attribute collision between cross-attention maps. 

\textbf{4.2 Reference color image Value matrices extraction} 

To perform image-guided color editing, we first need to convert the reference color image $I^c$ into its latent code $z^c_T$. And then extract the Value matrices $V^c$ of the cross-attention layer when we reconstruct the reference color image $I^c$.  In our paper, we primarily utilize Null-text Inversion \cite{mokady2023null} as the image inversion method. After inverting the reference color image $I^c$ , we denoise the latent codes $z^c_T$ and utilize the PyTorch register forward hook method \cite{paszke2019pytorch} to extract the value matrices $V^c$ from each layer of cross-attention of the denoising network in all timesteps.  It is worth noting that for one reference color,  the Value matrices $V^c$ and latent code $z^c_T$ need to be extracted only once and then can be applied repeatedly to different source images to change the color of the object. 

\textbf{4.3 Color Attribute Alignment and Object Structural Preservation} 

Based on our findings, we propose an image-guided method to modify the color of an object by using the Value matrices of a reference image $V^c$ to render the Value matrices of the target image  $V^T$ , which we call color attribute alignment. Those modified Value matrices are then used to calculate the spatial features for the next layer, thereby injecting color information into the denoising process. By aligning the Value matrices early in the generation process, when only the shape and contour of the object are established, we can resolve attribute information collisions between the cross-attention maps of the object and the color. Additionally, since no color term is introduced in the text prompt, the cross-attention leakage of the color attribute and the object is eliminated. To perform color attribute alignment, we normalize $V^{T}$ of the target image using $V^c$ of the reference color image with an adaptive normalization operation (AdaIN) \cite{huang2017arbitrary}:
\[
V^* =AdaIN(V^{T}, V^c), \quad (3) 
\]
where the AdaIN operation is given by:
\[
AdaIN(x, y) =\sigma(y)(\frac{x - \mu(x)}{\sigma(x)}) + \mu(y), \quad (4) 
\]
and $\mu(x), \sigma(x) \in R^{dk}$ are the mean and standard deviation of the value matrices (and the pipeline is given in Section 7.4).  However, the aligned value matrices may also introduce additional information, potentially changing the structure of the object.  To address this issue, we replace the self-attention map of the target image $M^{T}_{self}$ with the self-attention map of the source image $M^{S}_{self}$, since PnP \cite{tumanyan2023plug}, MasaCtrl \cite{cao2023masactrl}, and FreePromptEditing \cite{ liu2024towards} have shown that self-attention maps preserve spatial and structural information of an image.  So, our denoising process is given by 
\[
z^{T}_{t\text{-}1}= 
\begin{cases}
\text{VAlignDM}(z^{T}_t, P, t, V^c)  \{M^{T}_{self} \gets M^{S}_{self}\}& \text{$if \space t > \tau$} \\
\text{DM}(z^{T}_t, P, t)  \{M^{T}_{self} \gets M^{S}_{self}\} & \text{$otherwise$}
\end{cases}
\]
where $\text{VAlignDM}(z_t, P, t, V^c)$ is the alignment of the color attribute $AdaIN(V^{T}, V^c)$ in the cross-attention layer in the 1st, 2nd and 3rd CrossAttnUpBlocks of U-net.  And $\tau$ is a timestamp parameter that determines when we align the Value of cross-attention, a small figure of $\tau$ might lead to a higher broken structure of the object and a more change of the color of the object (see the quantitative experiment result in Section 8.4).   

\textbf{4.4 Object Segment, Latent blending, and Background Preservation} 

The objective of this study is to alter the object's color while maintaining the background region unchanged. To achieve this aim, it is necessary to segment the object from the background in the source image. In P2P\cite{hertz2022prompt} and MasaCtrl \cite{cao2023masactrl}, the binary mask was extracted directly from the cross-attention maps of the subject. However, the binary mask does not always precisely segment the object, especially around its edges. Observing that the cross-attention map from the 1st CrossAttnUpBlocks provides clear spatial location information of the object, we employ the state-of-the-art segmentation model SAM \cite{kirillov2023segment} to generate the mask of the object $Mask_{obj}$. More detailed information is given in Section 7.5.  To enhance the effectiveness of editing, we integrate the information from the latent code $z^c_T$ of the reference color image into $z^{s}_T$ with the object mask $Mask_{obj} $ at the beginning of the denoising process and call it latent blending. 
\[
 z^{T}_T =  (1 \text{-} Mask_{obj} ) * z^{s}_{T} \text{+} Mask_{obj} * ( z^{s}_T * (1 - R) \text{+} z^c_T * R ) 
\]
A higher ratio $R$ may enhance the color change, but could also cause the object to undergo more achromatic texture change (see the quantitative experiment result in Section 8.5) . In practice, we observe that setting the ratio to 0.1 yields the best quantitative results, while setting it to 0.15 gives better color changes perceived by humans. In order to preserve the background information, we update $z^{T}_t$ with the latent $z^{s}_t$ of the source image $I^s$ in the final few steps of the denoising process. 
\[
z^{T}_{t}= (1 \text{-} Mask_{obj} ) * z^{s}_t \text{+} Mask_{obj} * z^{T}_t 
\]
where  $z^{T}_t$ is the latent variable of the target image which is calculated after alignment of the value matrices and the replacement of the self-attention map. 
\section{Experiments}
\label{sec:experiments}
\begin{figure*}
    \centering
    \includegraphics[width=1.0\textwidth]{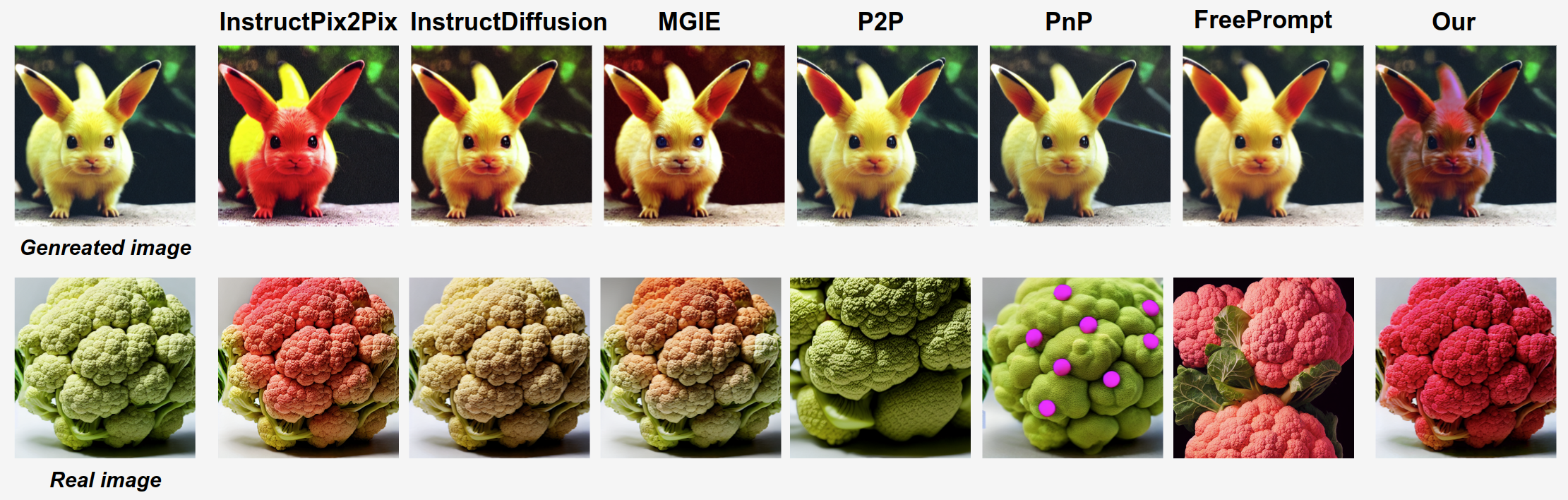}
    \caption{\textbf{Comparisons to text-guided editing methods.} For generated image, the text prompt is "A photo of pikachu" and the object is "pikachu". For real image, the object is "cauliflower". For real image and the P2P method,  we use Null text inversion method inverse the real image and then use P2P to change the color of the object. Here, we show the result of the color red. The result of other colors is presented in the Supplementary Material.} 
    \label{fig:color_editing_comparison_real_generated_image}
\end{figure*}

We implement the proposed method on the text-to-image Stable Diffusion model \cite{rombach2022high} with publicly available checkpoints v1.4.  During the sampling process, we used DDIM sampling \cite{song2020denoising} with 50 denoising steps and set the classifier-free guidance scale to 7.5.  We mainly compare our tune-free method with current state-of-the-art text-based diffusion editing methods, including training-based methods InstructPix2Pix \cite{brooks2023instructpix2pix}, Instructdiffusion \cite{geng2024instructdiffusion} and MGIE \cite{fu2023guiding}, fine-tuning-free approaches P2P \cite{hertz2022prompt}, PnP \cite{tumanyan2023plug} and FreePromptEditing \cite{liu2024towards}. We did not compare our method with the testing time fine-tuning method Imagic \cite{kawar2023imagic}, since it is not publicly available at this time.  For our method,  if not specifically indicated, the alignment of the cross-attention Value matrices is performed in the interval of $t \in [0.8T, T]$ of the denoising process, the reference color image latent blending ratio be 0.1, and background preservation at the last 5 timesteps. 

\textbf{5.1. Qualitative Experiments}

\begin{figure}
    \centering
    \includegraphics[width=0.5\textwidth]{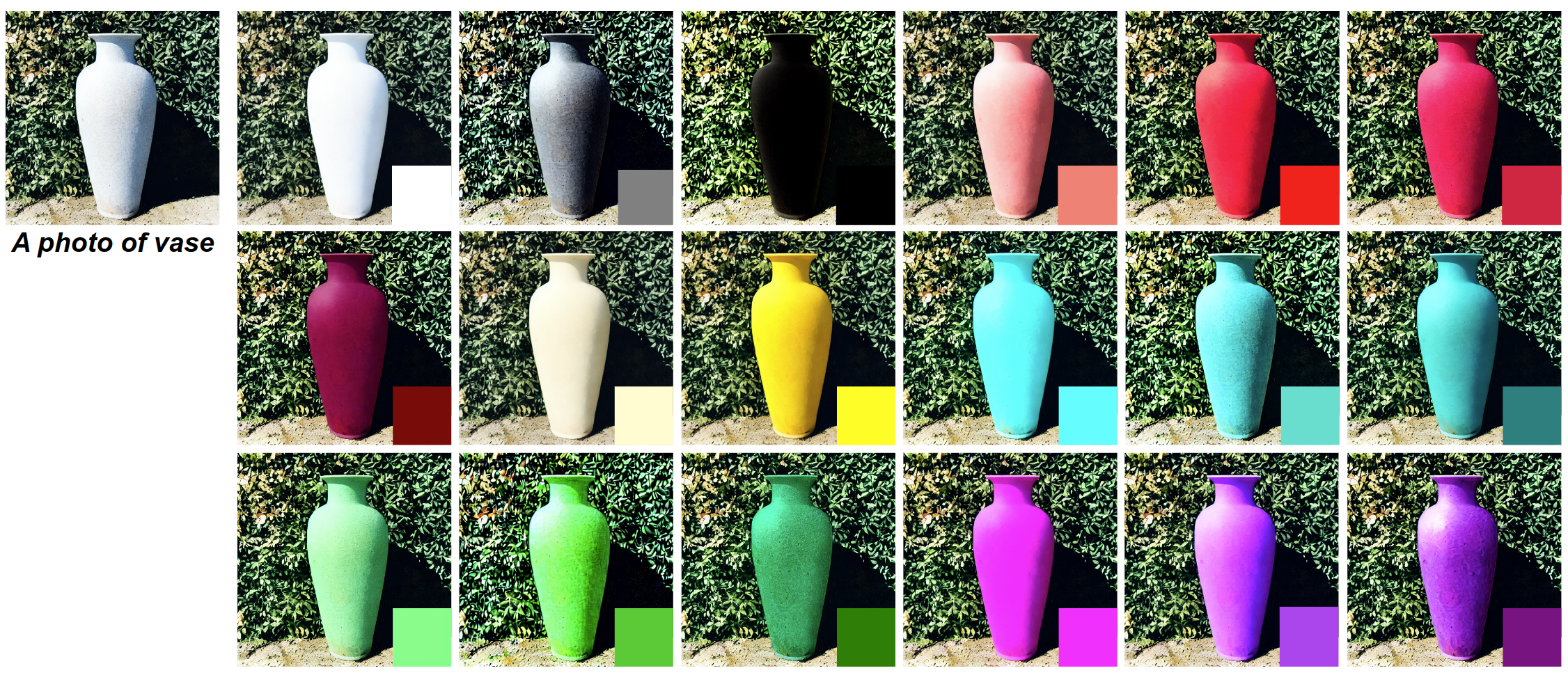}
    \caption{\textbf{Color editing utilize various reference images within the same color hue on generated image.} As demonstrated, our method enables more refined and controlled color changes.}
    \label{fig:visualing_result_across_various_color_in_the_color_red}
\end{figure}

\begin{figure}
    \centering
    \includegraphics[width=0.5\textwidth]{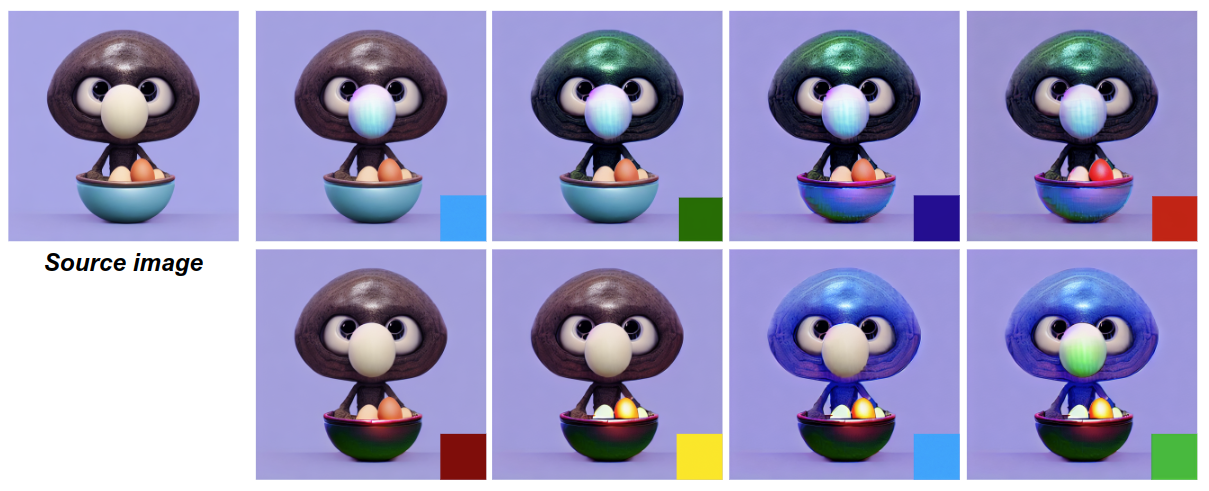}
    \caption{\textbf{Color editing with different sizes of object on real image.} As demonstrated, Our method can edit objects of different sizes in the same image while keeping preview results unchanged.}
    \label{fig:visualing_result_of_different_size_object}
\end{figure}

For generated images, as illustrated in the first row of Fig.~\ref{fig:color_editing_comparison_real_generated_image}, P2P and FreePromptEditing can preserve the background of the source image but fail to preserve the information around the object's contour (see the left ear of the \textit{pikachu}). All of those training-free methods may fail to edit the color of objects, especially when the object's color is uncommon or unlikely to occur in the real world. For training-based methods, InstructPix2Pix, Instructdiffusion, and MGIE can alter the color of the object. InstructPix2Pix can preserve the visual structure of the object while Instructdiffusion and MGIE might change the structure of the object. However, these training-based methods may inadvertently alter the background color of the target image. Our method is capable of modifying the color of the object while preserving both the object's structure and the background information. For real images, as demonstrated in the second row of Fig.~\ref{fig:color_editing_comparison_real_generated_image}, compared with training-free methods P2P, PnP, and FreePromptEditing, our method can preserve the structure of the object and change the color of the object. Furthermore, our method achieves the same or even better results compared to the results of training-based methods. Moreover, as shown in Fig.\ref{fig:visualing_result_across_various_color_in_the_color_red}, our method is capable of modifying the color of the object in a more refined and controlled way, and Fig.\ref{fig:visualing_result_of_different_size_object} indicates our method can edit different sizes of objects in the same image while keeping preview turn editing results.  For qualitative experiments, more results can be found in Section 8.1. 

\begin{table*}
    \centering
\begin{tabular}{c>{\centering\arraybackslash}p{0.03\linewidth}>{\centering\arraybackslash}p{0.06\linewidth}>{\centering\arraybackslash}p{0.06\linewidth}>{\centering\arraybackslash}p{0.06\linewidth}>{\centering\arraybackslash}p{0.06\linewidth}>{\centering\arraybackslash}p{0.06\linewidth}>{\centering\arraybackslash}p{0.06\linewidth}>{\centering\arraybackslash}p{0.09\linewidth}>{\centering\arraybackslash}p{0.09\linewidth}}
    \toprule
    & \multicolumn{6}{c}{\underline{\hspace{3.0cm}Generated dataset\hspace{3.0cm}}}& \multicolumn{3}{c}{\underline{\hspace{1.7cm}ColorBench\hspace{1.7cm}}}\\
 Method& $DS\uparrow$& $SSIM\uparrow$& $CS\uparrow$& $L1^{Hue}_{obj} \downarrow$& $L1^{HSV}_{obj} \downarrow$&$LPIPS_{bg} \downarrow$ & $SSIM\uparrow$& $LPIPS_{obj} \downarrow$&$LPIPS_{bg} \downarrow$\\
    \midrule
    P2P&  0.926& 0.803&  17.820&53.083& 58.851&0.042 & 0.6238& 0.332 &0.035\\
 PnP& 0.890& 0.762& 17.765& 54.837& 58.659&0.058 & 0.5557& 0.385&0.085\\
 FreePromptEditing& 0.911& 0.726& 17.897& 52.504& 58.619&0.077  & 0.6848& 0.265&0.016\\
 InstructPix2Pix&  0.949& 0.760& 17.711& \textbf{48.280}& 58.218&0.103  & 0.7750& \textbf{0.145}&0.044\\
 Instructdiffusion& 0.823& 0.673& 17.607& 50.849 & 58.327 &0.106 & 0.5517& 0.232&0.153\\
    MGIE&  0.89& 0.728&   17.698&53.395 & 60.624&0.08 & 0.6756& 0.208&0.076\\
    Ours& \textbf{0.957}& \textbf{0.83}&   \textbf{17.906}&50.800& \textbf{56.655}&\textbf{0.016} & \textbf{0.7751}& 0.195&\textbf{0.006}\\
      \bottomrule
\end{tabular}
  \caption{\textbf{Quantitative evaluation on the generated dataset and ColorBench dataset using various methods.} }
  \label{tab:t1}
\end{table*}

\begin{table*}
    \centering
    \begin{tabular}{ccccccc}
    \toprule
    Method& $DS\uparrow$&$SSIM\uparrow$&   $CS\uparrow$&$L1^{Hue}_{obj} \downarrow$& $L1^{HSV}_{obj} \downarrow$&$LPIPS_{bg} \downarrow$\\
    \midrule
    w/o self-attn. Replacement&  0.811 & 0.676 &  17.891&51.528 & 57.513 &0.024 \\
 w/o cross-attn. Alignment& \textbf{0.965}& \textbf{0.875}& 17.765& 52.677 & 59.509 &\textbf{0.007}\\
 w/o reference color image blending& 0.916 & 0.783 & 17.726& \textbf{50.768}& 57.435 &0.018 \\
    Ours& 0.957& 0.83&   \textbf{17.906}&50.800& \textbf{56.655}&0.016\\
      \bottomrule
\end{tabular}
 \caption{\textbf{Ablation study.} Our method achieves the optimal balance between preserving object structure and the desired color change.}
 \label{tab:t3}
\end{table*}

\begin{table}
    \centering
  \begin{tabular}{ccccc}
    \toprule
    Method& $OCE\uparrow$&$OSP\uparrow$&   $BP\uparrow$&$OEQ\uparrow$\\
    \midrule
    P2P
&  1.520 & 4.586&  4.932 &1.968
\\
 PnP
& 0.850 & 4.438& 4.688 &1.438
\\
 FreePromptEditing
& 1.344 & 4.248& 4.562 &1.72
\\
 InstructPix2Pix
& 3.468 & 4.744& 3.764 &3.03
\\
 Instructdiffusion
& 3.108 & 3.292& 3.920 & 2.284
\\
 MGIE
& 2.548 & 4.282& 4.318 & 2.474
\\
 Ours
&  \textbf{4.640}& \textbf{4.964}& \textbf{4.940}& \textbf{4.68}\\
      \bottomrule
\end{tabular}
  \caption{\textbf{Human evaluation results with different methods.} $OCE$ represent Object Color Editing. $OSP$ represent Object Structure Preservation. $BP$ represent Background Preservation. $OEQ$  represent Overall Editing Quality. }
  \label{tab:t2}
\end{table}

\textbf{5.2. Quantitative Experiments}

In the absence of publicly available datasets to validate the effectiveness of color change editing methods, we developed a generated dataset and a real image dataset for quantitative assessment. For the generated dataset, we selected 160 most common subjects from various categories with ChatGPT. For each subject, we automatically generated 7 prompts, resulting in 1,120 image-prompt pairs. For each source image, we used the SAM method \cite{kirillov2023segment} to generate the object mask and manually selected the best one. We then modified the object to seven different colors—white, gray, black, red, yellow, blue, and green—resulting in a total of 7,840 pairs of source and target images. For the real dataset, we manually crafted 406 images from 100 subjects with Photoshop, resulting in a total of 2,842 pairs of source and target images, which we call COLORBENCH. More information about the datasets is given in Section 8.2. For the color changing task, we separately evaluate three important aspects. First, we aim to achieve high semantic similarity and structural similarity between the source image and the target image. Second, we expect to achieve the desired color change for the object. Third, the background information in the target image should undergo minimal alteration.  

For the generated dataset, we utilize the DINO score \cite{caron2021emerging} on a gray scale to measure semantic similarity,  SSIM \cite{wang2004image} to measure structural similarity, CLIP Score (CS) \cite{radford2021learning} to measure the similarity between the target image and its corresponding text description , using L1 loss of Hue and HVS ($L1^{Hue}_{obj}$ and $L1^{HSV}_{obj}$) of the object area between the target image and the corresponding pure color image to measure how well the color of the object is changed, and use LPIPS \cite{zhang2018unreasonable} of the background area to measure the preservation of background information ($LPIPS_{bg}$). For more details, see Section 8.3. The quantitative result is presented in Table \ref{tab:t1} , our method outperforms others in preserving the object structure, background information, and semantic alignment between the target image and its corresponding text description. In terms of color hue alteration, our training-free approach performs comparably to training-based methods at the same numerical level, and our method has the lowest $L1^{HSV}_{obj}$ value. For the real image dataset,  we utilize the SSIM \cite{wang2004image} to measure structural similarity. Using $LPIPS_{obj}$ to measure the color change of the object and using $LPIPS_{bg}$ to measure the preservation of background information.  As can be seen in Table \ref{tab:t1}, our method performs best in preserving structure and background information but is slightly inferior in changing the object's color compared to the training-based method InstructPix2Pix.

\textbf{5.3. Human Evaluation}

We performed a human evaluation to assess the quality of color modifications across various methods. We randomly sampled 100 source-target examples and let 10 annotators rate these images. Participants are asked to rate the following aspects on a scale from 0 to 5: 1) effectiveness of color editing on the object, 2) preservation of the object's structure, 3) background preservation, and 4) overall editing quality.  The human evaluation results are listed in Table \ref{tab:t2}.  As can be seen, our method performs the best in all aspects, and the results are consistent with the quantitative experiments. For more detailed information, see Section 8.6.

\textbf{5.4. Ablation Study}

We conducted an ablation study to evaluate the effectiveness of our image-guided color editing method, highlighting the significance of each component. These components are (1) self-attention map replacement; (2) alignment of the value matrices in the cross-attention layer of the U-net decoder at the early stage of the denoising process; and (3) blending the latent of the reference color image. The results are presented in Table \ref{tab:t3}, and it is clear that without self-attention map replacement, the structure of the image is broken, leading to the lowest figures of $DS$ and $SSIM$. Without cross-attention alignment, the structure of the image is well preserved ($DS$ and $SSIM$ are highest) but resulting in the lowest change in the color of the object ($L1^{Hue}_{obj}$ and $L1^{HSV}_{obj}$ are the highest). Without the reference color image blending, the effect of color change improves a little, but with the scrubbing of the loss of structure preservation. Our method can achieve the best balance between the preservation of the structure and the change in the color of the object. A visualization ablation example is given in Section 8.7 of the Supplementary Material. 
\section{Conclusion}
\label{sec:conclusion}
In this work, we visualized the intermediate process of text-guided image generation and discovered that the shape, contour, and texture of an object are established in the U-Net decoder during the early denoising stage.  We identified that cross-attention leakage and attribute collision between the original object and color term are key factors to the failure of text-guided methods in the color change task. Based on those findings, we introduce a training-free image-guided method to edit the color of a subject through the color attribute alignment in the Value matrices in the cross-attention layer of the U-Net decoder in the early denoising stage. Furthermore, we introduce the COLORBENCH, the first benchmark to evaluate the color change task. Although simple in design, comprehensive qualitative and quantitative results demonstrate the effectiveness of our method. \textbf{Limitation and Future Work.} Our approach encounters certain limitations and will be addressed in our future work, which includes that the smaller the object, the harder it is to change the color of it, and the multi-object color change task has to be performed in a multi-turn way.

\clearpage
\maketitlesupplementary

\label{sec:supplymentary}
The Supplementary material of this paper consists of several sections that provide additional information and support for the main content.

\section{Additional information for Method Section}

\textbf{7.1 Framework pipeline and pseudocode for editing generated and real image} 

Algorithm ~\ref{alg: ALG1} and algorithm \ref{alg: ALG2} describe the pseudocode for editing generated and real images, respectively. We use Null-text Inversion \cite{mokady2023null} as the image inversion method to extract the value matrices of the reference color image $I^c$ and the latent space representation of the reference color image $I^c$ and the real image. 

\begin{algorithm}[!h]
\caption{Object-level Color Editing Algorithm for a generated image}
\label{alg: ALG1}

    \begin{flushleft}
        \hspace{0cm}  \textbf{Input:}: A source prompt $P$,  an interesting subject $O$ , a reference color image $I^c$\\
        \hspace{0cm}  \textbf{Output:}  A edited image $I^{T}$  \\
    \end{flushleft}
    
    \begin{algorithmic}[1]
        \State $z^{s}_T \in N (0, 1)$, a unit Gaussian random value sampled with random seed
       \State   $\{z^{s}_t\}^T_{t=0} , M^{S}_{cross}, M^{S}_{self}  \gets \text{DM}(z^{s}_T, P)$
        \State $Mask_{obj} \gets SAM(M^{S}_{cross}, O,   Generate(z^{S}_0)$
        \State $z^c_T \gets NullText - inv(I^c)$
        \State $V^c \gets LDM(z^c_T )$  $\triangleright$ with additional register forward hook method
             
        \For {$i = 1$ to $n$} 
            \State $z^{T}_T =  (1 \text{-} Mask_{obj} ) * z^{s}_{T} \text{+} Mask_{obj} * (z^{s}_T * (1 - ratio) \text{+}  z^c_T * ratio) $
            \For {$t = T, T-1, ..., 1$}           
                \If {$t > \tau $}
                    \State $z^{T}_{t-1} = \text{VAlignDM}(z^{T}_t, P, t, V^c) \{M^{T}_{self} \gets M^{S}_{self}\}$
                \Else
                    \State $z^{T}_{t-1} = \text{DM}(z^{T}_t, P, t) \{M^{T}_{self} \gets M^{S}_{self}\} $  
                \EndIf
    
                \If {$t < T-N $}
                    \State $z^{T}_{t-1}= (1 \text{-} Mask_{obj} ) * z^{s}_{t-1} \text{+} Mask_{obj} * z^{T}_{t-1} $
                 \EndIf
            \EndFor
        \State $I^{T}\gets Generate(z^{T}_0)$ 
        \State $z^{T}_T \gets NullText - Inv(I^{T}, P)$
        \EndFor
      
        \State \Return $I^{T}$
    
    \end{algorithmic}
\end{algorithm}

\begin{algorithm}[!h]
\caption{Object-level Color Editing Algorithm for a real image}
\label{alg: ALG2}

    \begin{flushleft}
        \hspace{0cm}  \textbf{input:}  A source image $I^{s}$, a text prompt $P$,  an interesting subject $O$ in the prompt $P$,  a reference color image $I^c$\\
        \hspace{0cm}  \textbf{Output:}  A edited image $I^{T}$  \\
    \end{flushleft}
    
    \begin{algorithmic}[1]
        \State $\{z^{s}_t\}^T_{t=0}, M^{S}_{cross}, M^{S}_{self} \gets NullText - inv(I^{s}, P)$
        \State $Mask_{obj} \gets SAM(M^{S}_{cross}, O, I^{s})$
        \State $z^c_T \gets NullText - inv(I^c)$
        \State $V^c \gets LDM(z^c_T )$  $\triangleright$ with additional register forward hook method
                     
        \For {$i = 1$ to $n$} 
          \State $z^{T}_T =  (1 \text{-} Mask_{obj} ) * z^{s}_{T} \text{+} Mask_{obj} * (z^{s}_T * (1 - ratio) \text{+}  z^c_T * ratio) $
          \For {$t = T, T-1, ..., 1$}
                \If {$t > \tau $}
                    \State $z^{T}_{t-1} = \text{VAlignDM}(z^{T}_t, P, t, V^c) \{M^{T}_{self} \gets M^{S}_{self}\}$
                \Else
                    \State $z^{T}_{t-1} = \text{DM}(z^{T}_t, P, t) \{M^{T}_{self} \gets M^{S}_{self}\} $  
                \EndIf
                
                 \If {$t < T-N $}
                    \State $z^{T}_{t-1}= (1 \text{-} Mask_{obj} ) * z^{s}_{t-1} \text{+} Mask_{obj} * z^{T}_{t-1} $
                 \EndIf
                 
            \EndFor
        \State $I_{T} \gets Generate(z^{T}_0)$ 
        \State $z^{T}_T \gets NullText - Inv(I^{T}, P)$
        \EndFor
    
        \State \Return $I^{T} $
    
    \end{algorithmic}
\end{algorithm}

\textbf{7.2 Cross-attention map in different blocks of U-Net} 

Fig.~\ref{fig:cross-attention-map-of-object-different-block} 
visualizes the average cross-attention maps of various objects from different blocks across all time steps. The cross-attention maps from different blocks are resized to the same resolution (512×512). As observed, the U-Net encoder captures some structural information of the object, but it is blurred (see the attention map of the 2nd and 3rd CrossAttnDownBlocks). In contrast, the U-Net decoder provides a clearer representation of the object's shape, contour, and texture. Specifically, the 1st, 2nd, and 3rd CrossAttnUpBlocks of the U-Net reveal progressively clearer shape, contour, and texture, respectively.  

\begin{figure*}
    \centering
    \includegraphics[width=1.0\textwidth]{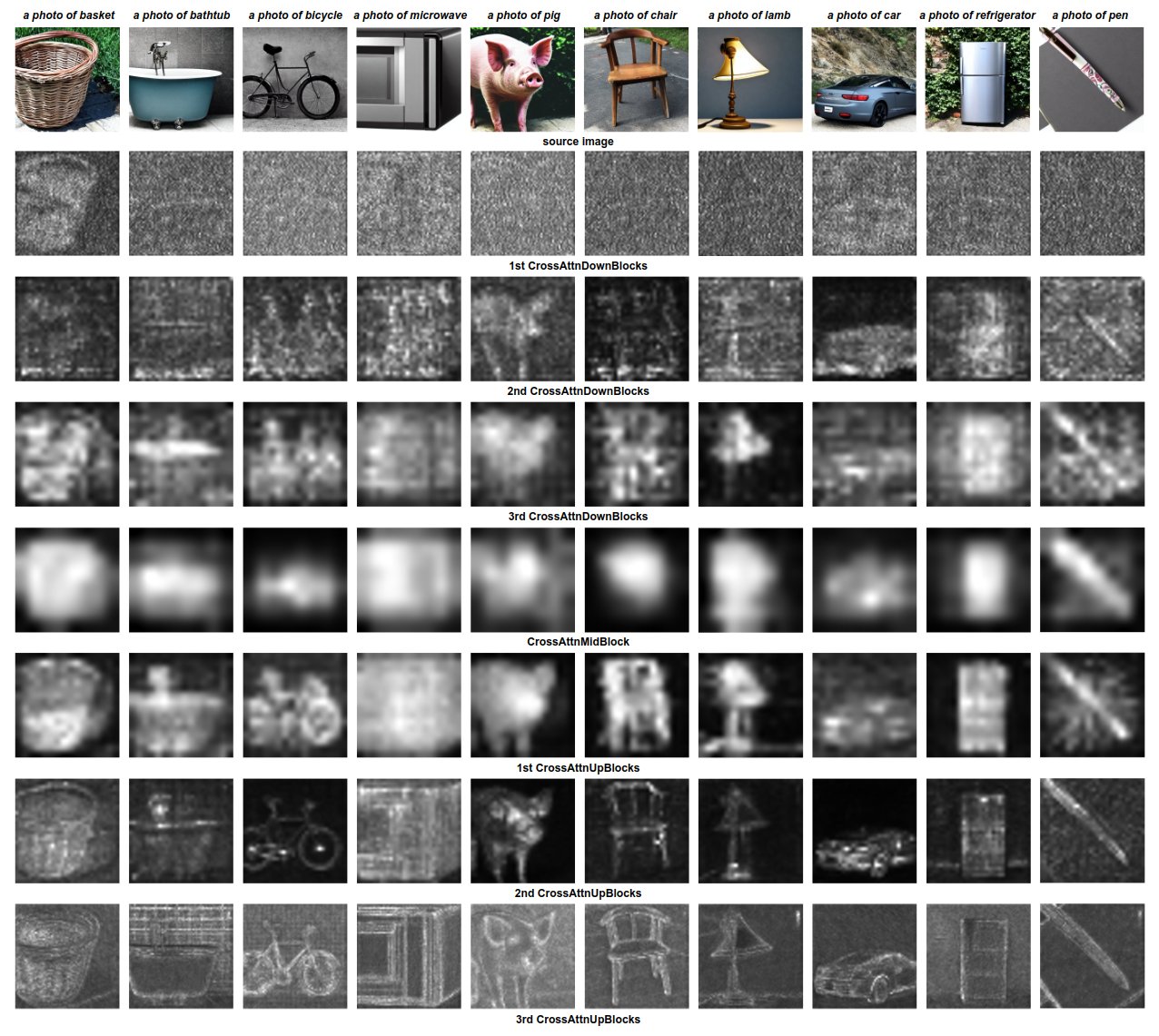}
    \caption{\textbf{Visualizing cross-attention map in the text-guided diffusion image generation.}  In this case, the text prompt is "a photo of *", where "*" represents a specific object. }    \label{fig:cross-attention-map-of-object-different-block}
\end{figure*}

\begin{figure}
    \centering
    \includegraphics[width=0.5\textwidth]{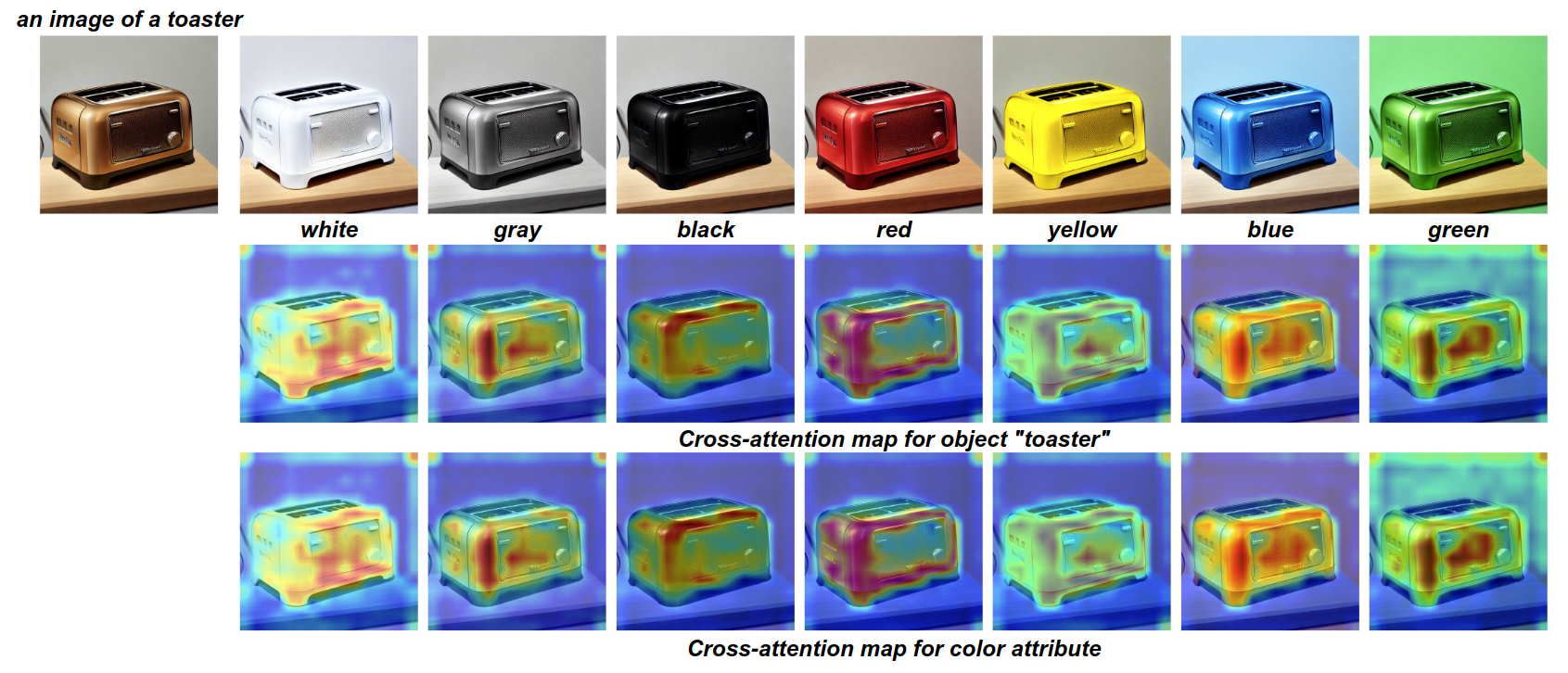}
    \caption{\textbf{Visualization of the Cross-attention maps for object and color attributes using InstructPix2Pix to edit the color. } In this example, the source image is generated with the prompt "an image of a toaster", while the edit text prompt is "turn the color of toaster to *," where "*" represents a specific color.} 
    \label{fig:instructPix2Pix_cross_attention_map_object_color}
\end{figure}
\textbf{7.3  Cross-attention leakage in training-based text-guided image editing methods} 

For training-based text-guided image editing methods \cite{brooks2023instructpix2pix, geng2024instructdiffusion, fu2023guiding}, the color change task is performed directly through text prompts, and the phenomenon of attention leakage also exists. As demonstrated in Fig.~\ref{fig:instructPix2Pix_cross_attention_map_object_color}, when the cross-attention map of the color attribute is distributed to the background of the image, the background color changes (e.g. when attempting to change the color of the toaster to white, gray, or green).

\begin{figure}
    \centering
    \includegraphics[width=0.5\textwidth]{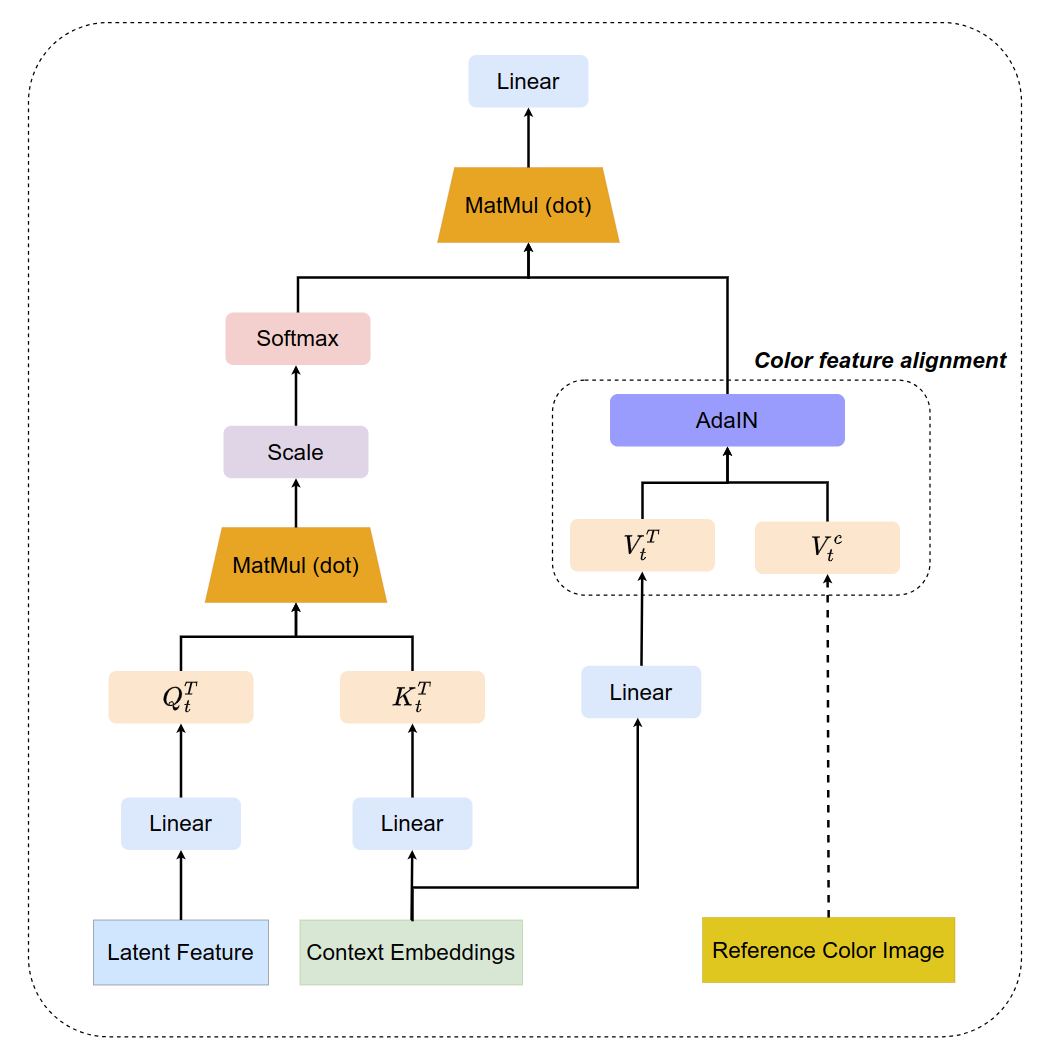}
    \caption{\textbf{Color attribute alignment in the cross-attention layer.}  The color of the object in the target image attends to the reference color image by applying AdaIN over its Value matrices $V^T$ using the Value matrices $V^c$ from the reference color image.  } 
    \label{fig:color_attribute_alignment_in_cross_layer}
\end{figure}

\textbf{7.4  color attribute alignment Pipeline} 

Fig.~\ref{fig:color_attribute_alignment_in_cross_layer} shows the color attribute alignment pipeline, color modification is achieved by utilizing the Value matrices of a reference image to normalize the Value matrices of the target image.
\begin{figure}
    \centering
    \includegraphics[width=0.5\textwidth]{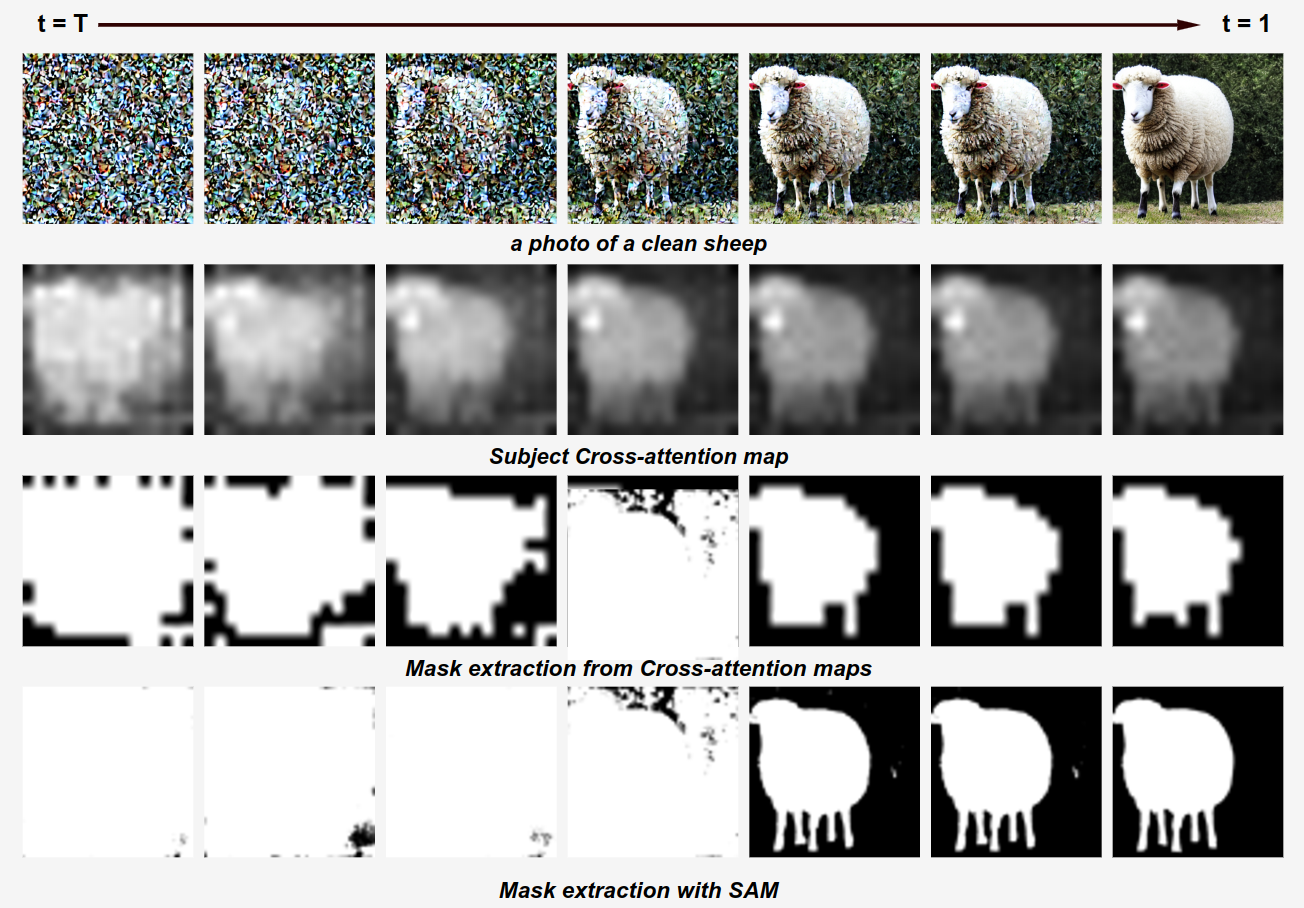}
    \caption{\textbf{Visualization of subject mask extraction using cross-attention maps in P2P and SAM. }\textnormal{In this example, all masks are resized to a resolution of $64 \times 64$ , and the text prompt is "a photo of a clean sheep".} }
    \label{fig:subject_mask_in_diferent_timestep}
\end{figure}

\begin{figure}
    \centering
    \includegraphics[width=0.5\textwidth]{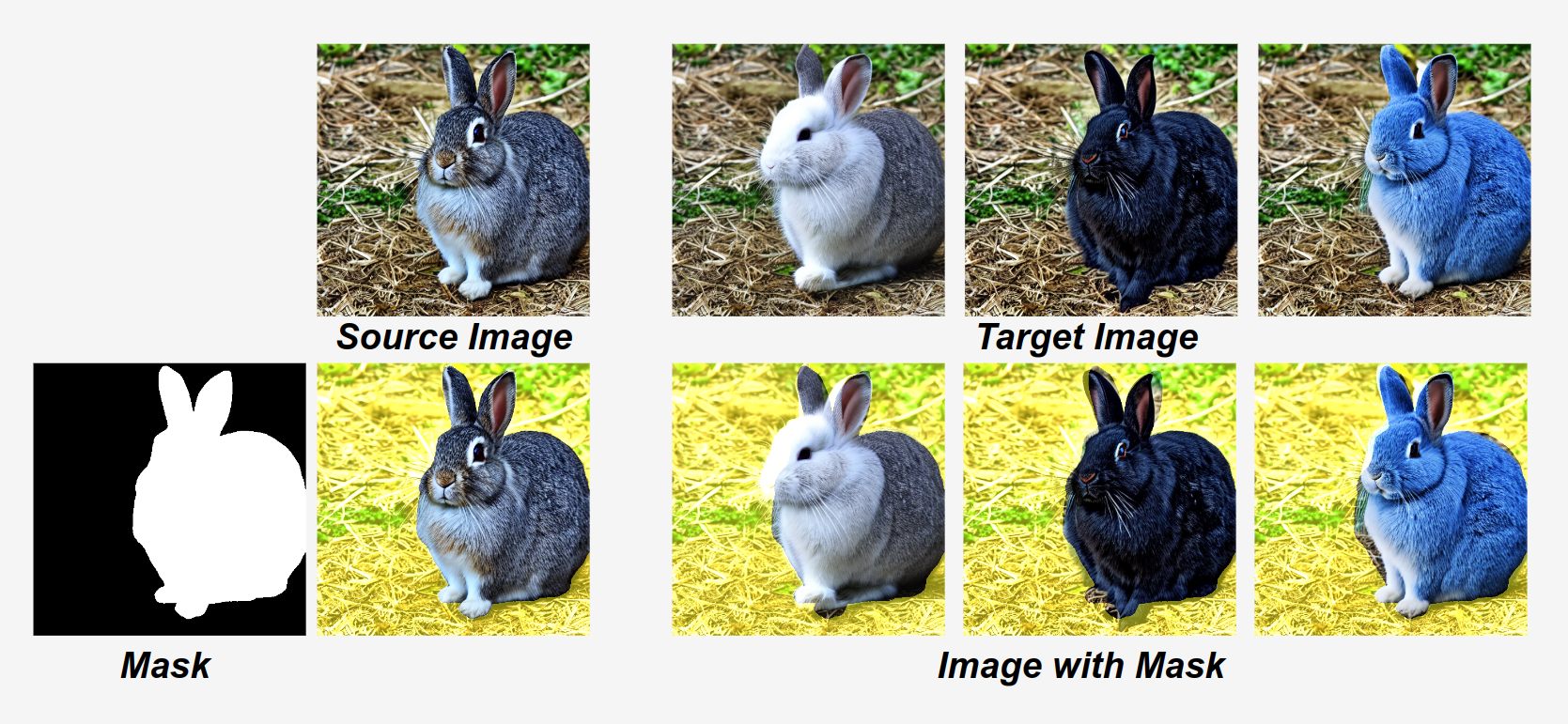}
    \caption{\textbf{Visualization of the edited image using P2P with the ground-truth mask.} As demonstrated, utilizing cross-attention maps to extract the binary mask of an object does not perfectly segment the object from the background, resulting in changes along the object's edges. In this example, the text prompt is "a photo of rabbit," and the target image, from left to right, shows the rabbit's color changing to white, black, and blue.} 
    \label{fig:subject_mask_with_cross_attention_map}
\end{figure}

\textbf{7.5 Object Mask generation and selection} 

In P2P\cite{hertz2022prompt} and MasaCtrl \cite{cao2023masactrl}, the binary mask was extracted directly from the cross-attention maps of the subject with a resolution of $16 \times 16$, which is obtained from the 3rd CrossAttnDownBlock and the 1st CrossAttnUpBlock in the U-Net. These cross-attention maps were then averaged and compared with a fixed threshold to generate the binary mask.  As seen in Fig.~\ref{fig:cross-attention-map-of-object-different-block}, the edges of the cross-attention maps in the 3rd CrossAttnDownBlock are blurred, leading to a binary mask that does not always precisely segment the object, especially around its edges (see the 3rd row in the Fig.~\ref{fig:subject_mask_in_diferent_timestep}) which resulting in changes along the object's edges ( see the example in Fig.~\ref{fig:subject_mask_with_cross_attention_map}, when we try to change the color of the rabbit to blue).

SAM \cite{kirillov2023segment} requires a segmentation prompt (e.g., point prompt) as input and generates a series of object masks $\{Mask_{obj}\}_n$. The point prompt (subject centroid) can be calculated from the cross-attention mask  $Mask^{cross}_{obj}$, which is obtained from the 1st CrossAttnUpBlock of the U-Net. To select the best object $Mask_{obj}$, we calculate a series of scores and choose the mask with $arg \space min_ns(i, n)$:
\[
s(i, n) = abs(1 - \frac{sum(Mask^{cross}_{obj})}{sum(Mask^{i}_{obj})} )
\]
where $sum(Mask)$ represents the area of zero value of the cross-attention mask or the mask generated from SAM. 

\begin{figure*}
    \centering
    \includegraphics[width=1.0\textwidth]{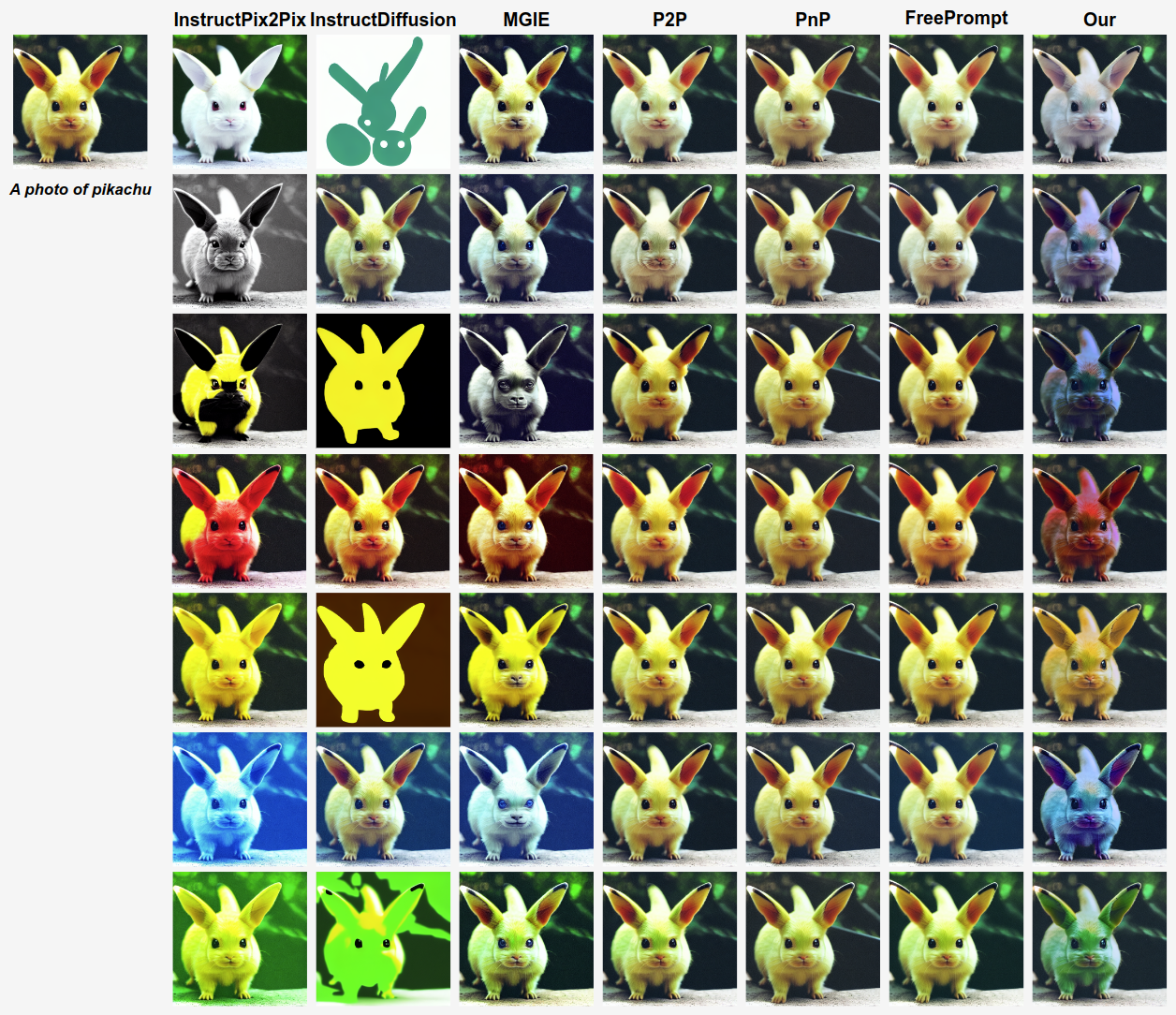}
    \caption{\textbf{Comparisons to text-guided editing methods for generated image.} In each row, we present the modified image reflecting the desired color for the object using various methods. From top to bottom, the desired colors are "white," "gray," "black," "red," "yellow," "blue" and "green. In this case, the text prompt is "A photo of pikachu" and the object is "pikachu".
  } 
    \label{fig:color_editing_comparison_geneartion_image_all_color}
\end{figure*} 

\begin{figure*}
    \centering
    \includegraphics[width=1.0\textwidth]{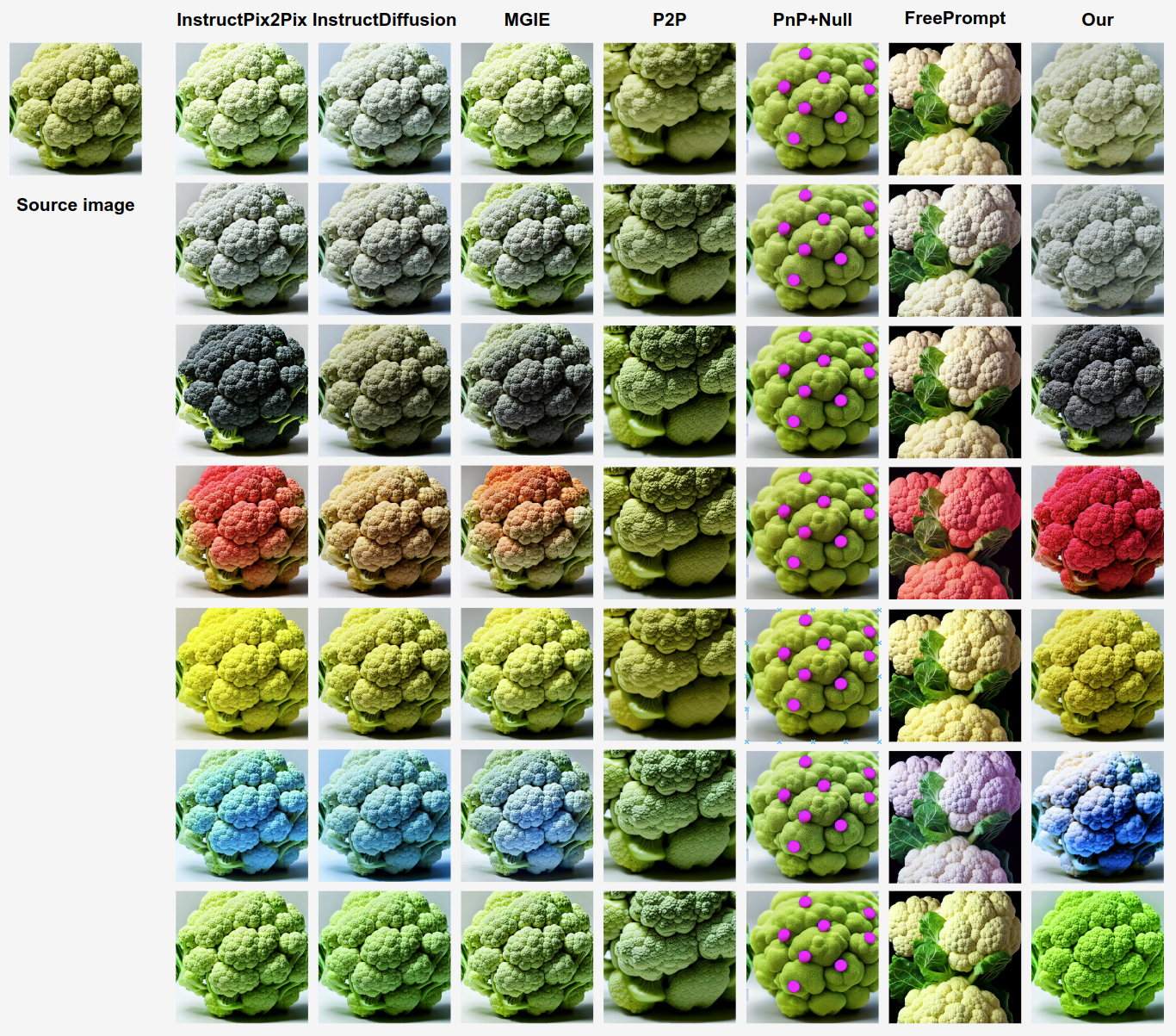}
    \caption{\textbf{Comparisons to text-guided editing methods for real images.} In each row, we present the modified image reflecting the desired color for the object using various methods. From top to bottom, the desired colors are "white," "gray," "black," "red," "yellow," "blue" and "green.
  } 
    \label{fig:color_editing_comparison_real_image_all_color}
\end{figure*}

\begin{figure*}
    \centering
    \includegraphics[width=1.0\textwidth]{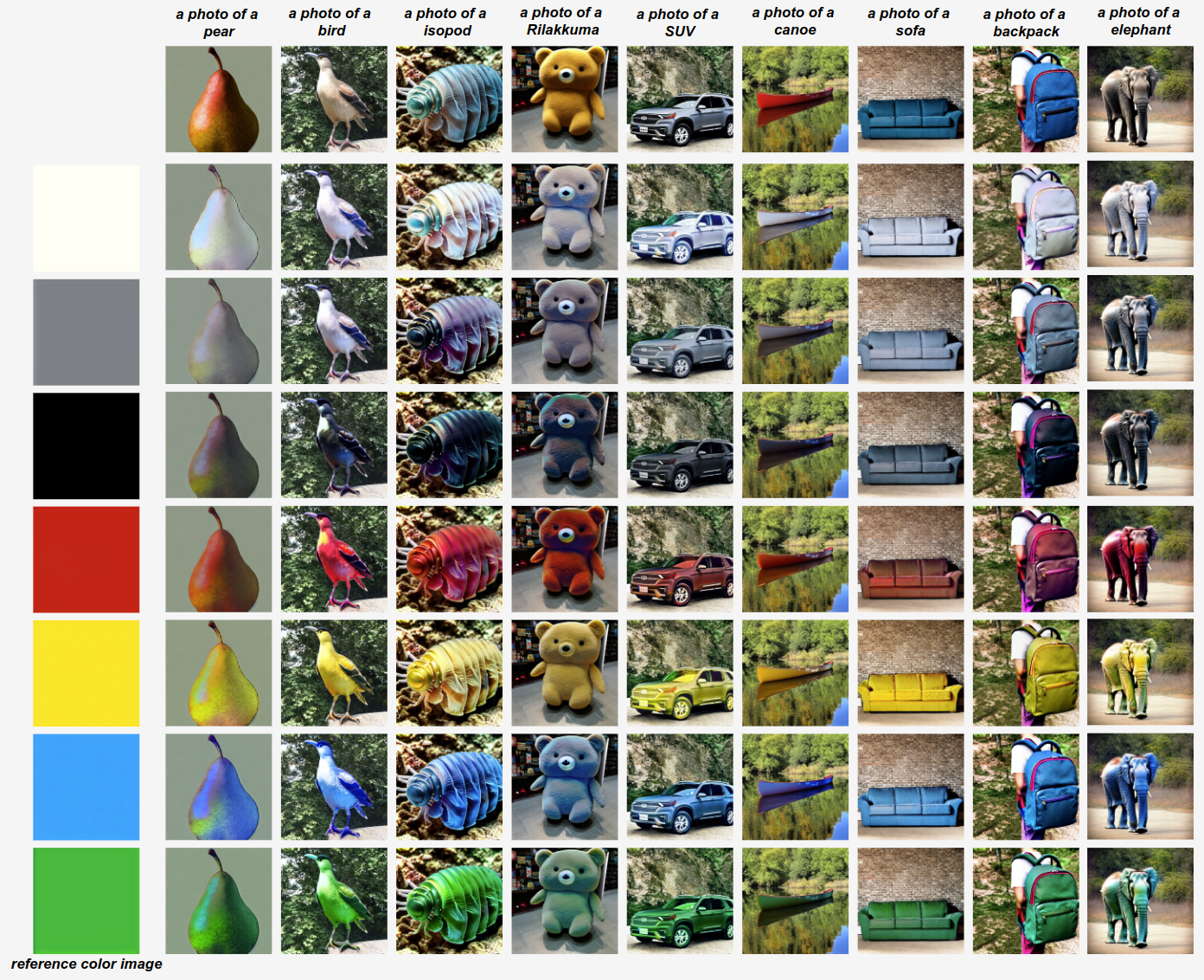}
    \caption{\textbf{Color editing for generated images.} The first row displays the source image, while the subsequent rows depict the modified image with the desired color.} 
    \label{fig:colorEditGenerateImage}
\end{figure*}

\begin{figure*}
    \centering
    \includegraphics[width=1.0\textwidth]{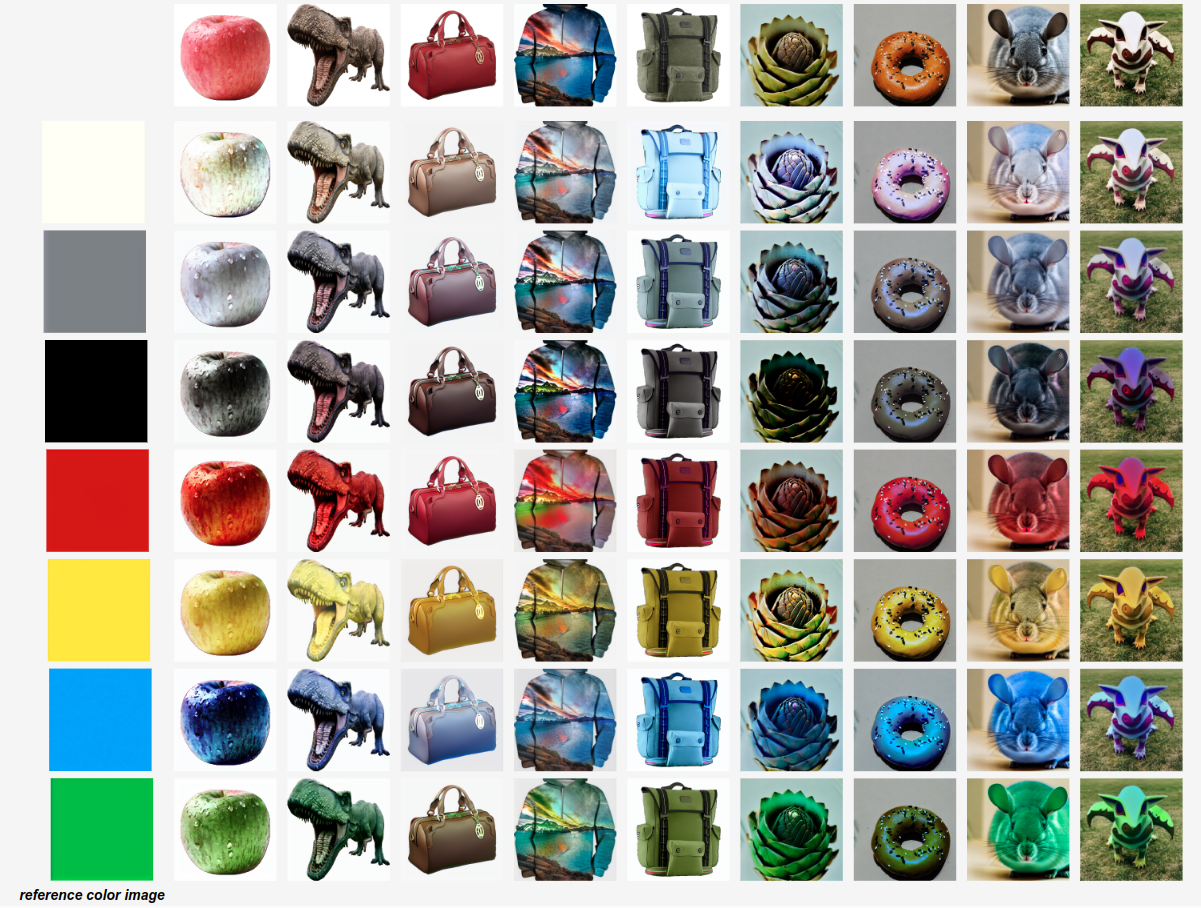}
    \caption{\textbf{Color editing for real images.} The first row displays the source image, while the subsequent rows depict the modified image with the desired color.} 
    \label{fig:colorEditrealImage}
\end{figure*}

\section{Additional information for Experiment Section}

\textbf{8.1 Qualitative Experiments} 

For symthesised images, we compare our method with text-guided image editing methods InstructPix2Pix \cite{brooks2023instructpix2pix}, Instructdiffusion \cite{geng2024instructdiffusion},  MGIE \cite{fu2023guiding} , P2P \cite{hertz2022prompt}, PnP \cite{tumanyan2023plug} and FreePromptEditing \cite{liu2024towards}. For training-based methods InstructPix2Pix \cite{brooks2023instructpix2pix}, Instructdiffusion \cite{geng2024instructdiffusion} and MGIE \cite{fu2023guiding}, the prompt is "turn the color of the pikachu to $*$" and $*$ represents the specified color. For fine-tuning free approaches, P2P \cite{hertz2022prompt}, PnP \cite{tumanyan2023plug}, and FreePromptEditing \cite{liu2024towards}, the target prompt is "a photo of * pikachu", where "*" is the specific color.  Fig.~\ref{fig:color_editing_comparison_geneartion_image_all_color} provides all the color change outcomes for the generated image, including white, gray, black, red, yellow, blue, and green.

For real images, we conduct a comparison of our method with the text-guided image editing method InstructPix2Pix \cite{brooks2023instructpix2pix}, Instructdiffusion \cite{geng2024instructdiffusion},   MGIE \cite{fu2023guiding},  P2P \cite{hertz2022prompt}, PnP \cite{tumanyan2023plug} and FreePromptEditing \cite{liu2024towards}.  Since P2P cannot edit the real image, we use the Null text inversion \cite{mokady2023null} method to inverse the real image to its latent variable and then use P2P to change the color of the object.  Fig.~\ref{fig:color_editing_comparison_real_image_all_color} provides all the color change results for real images, including white, gray, black, red, yellow, blue, and green.

Fig.~\ref{fig:colorEditGenerateImage}  and Fig.~\ref{fig:colorEditrealImage} show the result of the generated image and the real image of various objects and image domains, including fruit, bird, insect, furry doll, plant, food, animal, toy, etc. Notice the color change of the object and the preserved background of the source image.  

\begin{figure*}
    \centering
    \includegraphics[width=1.0\textwidth]{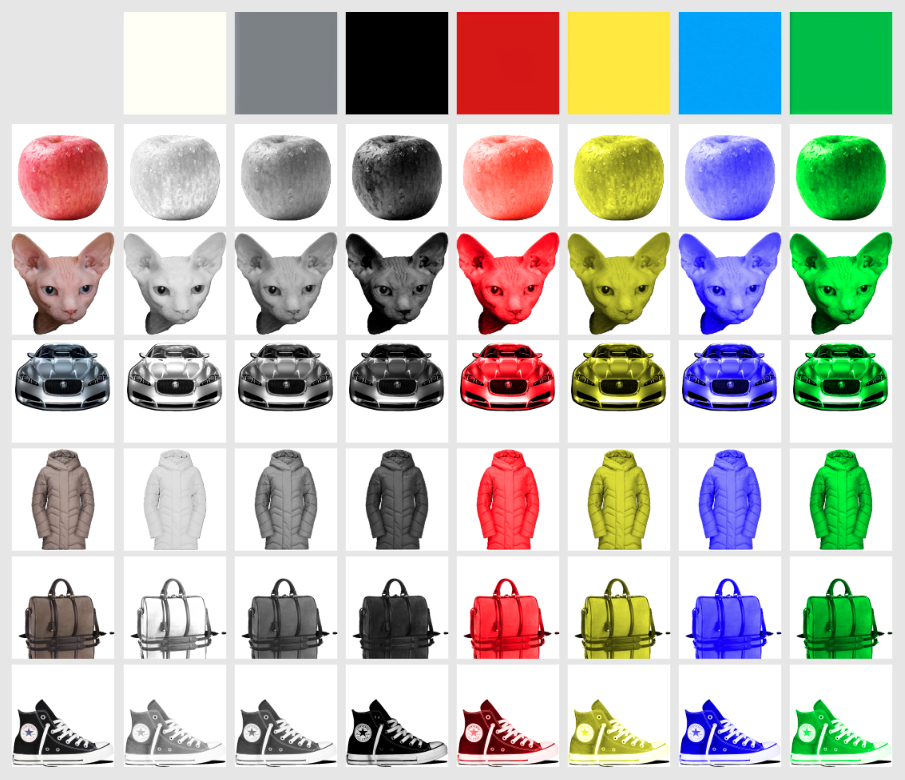}
    \caption{\textbf{Example data of COLORBENCH.} The COLORBENCH dataset including 406 images from 100 subjects, and the color of object in each image are change to 7 different colors, including white, gray, black, red, yellow, blue, and green.} 
    \label{fig:colorBenchSample}
\end{figure*}

\textbf{8.2 Data collection for Quantitative Experiments} 

For the generated dataset, the category of the 160 subjects includes fruits, vegetables, animals, insects, birds, musical instruments, furniture, electronic products, clothing, food, vehicles and cartoon characters. The list of the subjects are bread, hedgehog, raccoon, van, panda, pitcher, baseball, Pikachu, perch, helmet, butterfly, spinach, sofa, kale, onion, tarantula, minivan, chinchilla, Charmander, chair, soap, sunglasses, dolphin, porcupine, hippopotamus, kinkajou, ant, pineapple, hamburger, bulb, honeydew, fox, piranha, bison, donut, finch, lizard, Rilakkuma, waffle, cattle, mushroom, bagel, bench, python, tomato, tie, boots, leek, bongo, giraffe, bed, bow, hovercraft, scooter, wolf, bee, egg, sweater, dog, cymbal, lettuce, starfish, sparrow, starfruit, celery, turtle, isopod, lime, humidifier, parrot, ladle, armadillo, monkey, avocado, snake, skunk, bird, toaster, rat, koala, camel, apricot, tank, apple, hoodie, squirrel, lemon, tiger, cantaloupe, weasel, sheep, rabbit, snail, elephant, bear, Pichu, hamster, goldfish, coat, shorts, ladybug, lion, ball, gun, SUV, otter, tick, table, beaver, mackerel, gerbil, handbag, artichoke, asparagus, cake, cauliflower, speaker, ocelot, banana, backpack, broccoli, Cubone, lemur, scorpion, javelina, canoe, timpani, dragonfly, cup, pumpkin, coconut, iguana, lobster, orange, snowman, deer, parsnip, peach, limousine, pear, purse, ferret, antelope, spoon, lamp, pillow, carrot, cabbage, truck, frog, grapefruit, drum, brownie, moose, Simba, scarf, mango, glider, cello.  The text templates we use to generate the prompt are: 1) "a photo of a \{\}", 2) "an image of a \{\}" , 3) "a photo of a nice \{\}", 4) "a photo of a large \{\}", 5) "a good photo of a \{\}", 6) "a rendition of a \{\}", and 7) "a toy of a \{\}". For quantitative analysis, we generated a dataset comprising 1,120 image-prompt pairs, with the source images created with random seed.  

For real image dataset, the dataset includes 406 images from 100 subjects, which include: apple, cat, dinosaur, garlic, ladybug, peach, strawberry, asparagus, celery, dog, glove, leaf, peanut, sugar, avocado, champagne, doughnut, goldfish, lemon, pear, suitcase, backpack, cherry, dragon, giraffe, lion, pigeon, tomato, banana, chick, dragonfly, grape, macaron, potato,  T-shirt,  bee,  chili,  eagle,  hamburger,  mango, raspberry,  bird,  chiwawa,  earthworm, handbag, mantis, rhino, vegetable, blueberry, coat, eggplant, hat,  mushroom,  rock, walnut,  boots, cocktail, elephant,   heel,  noodle, sausage, wasp, bug, coconut, fig,   hornet, nut, scorpion, watermelon, butter, cookie, fish, horse, okra, Shells, yam, butterfly, cricket, flea, icecream, orange, shirt, cantaloupe, Croton, flower, insect, pancake, shoe, cap, cucumber, fly, kiwifruit, parrot, spider, car, daikon, fruit, ladybird, pea, spinach. To remove the influence of background information, we set the background to white color. An example of COLORBENCH is given in Fig.~\ref{fig:colorBenchSample} .

\textbf{8.3 Evaluation Metrics for Quantitative Experiments} 

For evaluation metrics, we measure semantic similarity using the DINO score \cite{caron2021emerging}, which is designed to assess how similar two images are in terms of their semantic content, including objects, scenes, and concepts. However, to eliminate the effect of the object's color on the DINO score, we compute the DINO score using the grayscale versions of both the source and target images and call it $DS$.  For structural similarity, we applied SSIM \cite{wang2004image}, which evaluates the similarity between two images based on luminance, contrast, and structure at the pixel level.   To verify that the edited image contains the specified color of the object, we calculate the CLIP Score (CLIP cosine similarity) \cite{radford2021learning} between the image and its corresponding text description. For instance, if the source image is generated with the prompt "a photo of a squirrel" and the color of the squirrel is altered to red, the CLIP cosine similarity is computed between the edited image and the text description "a photo of a red squirrel". A higher CLIP score (CS) indicates that the image and text are semantically aligned, which means that the image accurately depicts the content described by the text.  Although the CLIP score effectively measures semantic alignment, it does not directly evaluate other critical aspects of image quality, such as texture, color accuracy, or structural fidelity. To assess color accuracy, we calculate the L1 loss between the target image and the corresponding pure color image within the object area using the HSV color system, since the HSV color system provides a more human-friendly way of manipulating colors compared to the RGB model, which we call $L1^{HSV}_{obj}$.  To eliminate the influence of color purity and lightness, we exclude the saturation and value components and calculate the L1 loss based solely on the hue, which we call the loss L1 of Hue of the object $L1^{Hue}_{obj}$. A lower L1 loss signifies a closer alignment between the object and the desired target color hue. To quantify the preservation of background information, we employ the object's mask to isolate the background in both the source and target images, subsequently evaluating the similarity using LPIPS \cite{zhang2018unreasonable} and call it $LPIPS_{bg}$.  A lower LPIPS value indicates better preservation of background information. 

For color accuracy measure, the RGB values we use to construct the pure color image are: 1) black: (0, 0, 0), 2) white: (255, 255, 255), 3) gray: (128, 128, 128), 4) red: (255, 0, 0), 5) yellow: (255, 255, 0), 6) blue: (0, 0, 255), 7) green: (0, 255, 0).

\begin{figure*}
    \centering
    \includegraphics[width=1.0\textwidth]{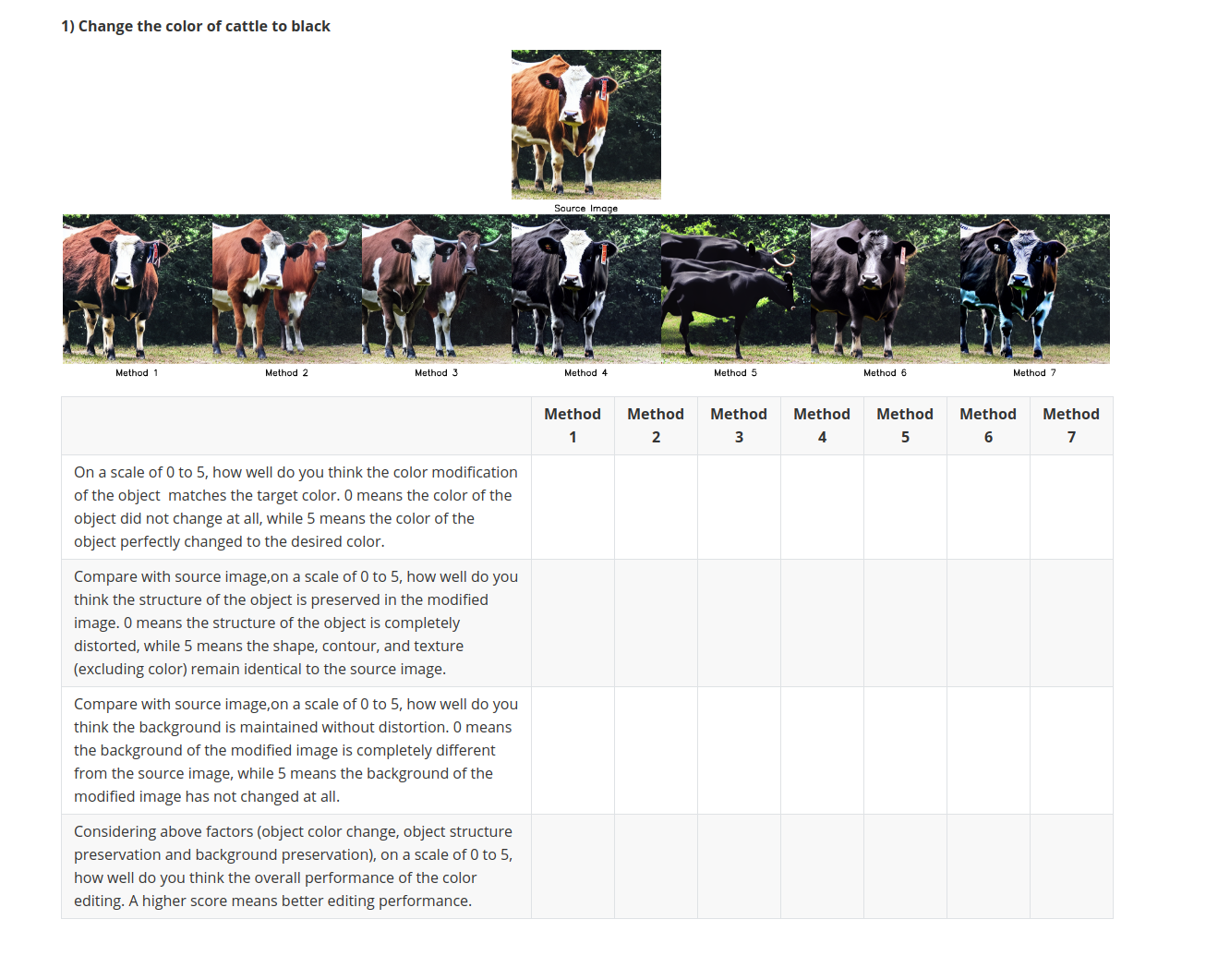}
\caption{\textbf{Screenshot from the Human Evaluation.} The second row of the image shows the results obtained by different methods. The user is asked to rank the performance of each method from 0 to 5. 0 means the worst,while 5 means the best. }
  
    \label{fig:human_evaluation_example}
\end{figure*}

\begin{table*}
    \centering
  \begin{tabular}{ccccccc}
    \toprule
    & $DS\uparrow$&$SSIM\uparrow$&   $CS\uparrow$&$L1^{Hue}_{obj} \downarrow$& $L1^{HSV}_{obj} \downarrow$&$LPIPS_{bg} \downarrow$\\
    \midrule
    w/o cross-attn. Alignment&  \textbf{0.965}& \textbf{0.875}&  17.766 &52.677 & 59.509 &\textbf{0.007}\\
 First 20\% time step
(our)& 0.957 & 0.830 & 17.906 & \textbf{50.800}& 56.655 &0.016 
\\
 First 50\% time step& 0.876 & 0.715 & \textbf{17.955}& 50.864 & \textbf{55.944}&0.032 
\\
    All time step& 0.622 & 0.622 &   17.943 &52.084 & 55.962 &0.046 
\\
      \bottomrule
\end{tabular}
  \caption{\textbf{Quantitative evaluation results with cross-attention value alignment in different diffusion time step.} The reference color image blending ratio is set to 0.1}
  \label{tab:t4}
\end{table*}

\textbf{8.4 Quantitative evaluation results for Cross-attention value Alignment} 

 Table \ref{tab:t4} shows that without alignment of cross-attention values, the structures of the object remain well preserved. However, as cross-attention values are aligned during further denoising steps, the object's structural integrity progressively deteriorates (see how $DS$ and $SSIM$ change through different cross-attention value alignment plans). For the color change of the object, the effect improves as the number of alignment steps of values increases. However, it decreases when the alignment of the cross-attention value is applied throughout the entire denoising process (see how $L1^{Hue}_{obj}$ and $L1^{HSV}_{obj}$ change in the table).

\begin{table*}
    \centering
  \begin{tabular}{ccccccc}
    \toprule
    Ratio& $DS\uparrow$&$SSIM\uparrow$&   $CS\uparrow$&$L1^{Hue}_{obj} \downarrow$& $L1^{HSV}_{obj} \downarrow$&$LPIPS_{bg} \downarrow$\\
    \midrule
    0.0&  0.916 & 0.783 &  17.726&50.768& 57.435 &0.018 \\
 0.05& 0.948 & 0.829 & 17.798 & 51.092 & 57.649 &\textbf{0.015}\\
 0.1& \textbf{0.957}& \textbf{0.830}& 17.906& 50.800& \textbf{56.655}&0.016\\
    0.15& 0.946 & 0.795 &   \textbf{17.925}&\textbf{50.721}& 56.935 &0.018 
\\
 0.2& 0.922 & 0.760 & 17.862 & 50.860 & 57.944 &0.020 
\\
      \bottomrule
\end{tabular}
  \caption{\textbf{Quantitative evaluation results  with different reference color image latent blending ratio.} The alignment of the attention value is performed in the first 20\% of the diffusion time step.} 
    \label{tab:t5}
\end{table*}

\textbf{8.5 Quantitative evaluation results for different reference color image blending ratio} 

 Table \ref{tab:t5} shows the quantitative evaluation result of different latent variable blending ratios of the reference color image. As shown in the table, setting the ratio to 0.1 results in the least structural distortion of the object. However, when the ratio is set to 0.15, the color change effect is optimal, though there is a slight loss in the object's structural integrity.

\textbf{8.6 Details of Human Evaluation} 

As described in the main paper, we randomly select 100 pairs of source and target images.  The distribution of desired color in target images is: 15 white, 14 gray, 13 black, 14 red, 15 yellow, 14 blue, and 15 green. The involved subjects include: cattle, celery, handbag, honeydew, onion, perch, waffle, drum, orange, pineapple, tomato, finch, spinach, artichoke, banana, cup, coconut, dragonfly, lizard, porcupine, egg, ferret, lime, wolf, fox, mushroom, snowman, pumpkin, bread, lettuce, ladybug, tarantula, lobster, Pichu, bench, bongo, butterfly, cello, coat, hoodie, melmet, isopod, baseball, pillow, dolphin, lion, tank, rat, speaker, kale, lemon, tiger, asparagus, Pikachu, grapefruit, javelina, sofa, starfruit, bagel, piranha, and apple.

The color editing effect of an object refers to how closely the object's color modification aligns with the target color. Object identity preservation measures the extent to which the object's original identity is retained in the edited image. The preservation of the background indicates how well the background remains intact without distortion. Overall editing quality encompasses the naturalness of the color modification, including the overall realism of the image post-editing.

The human evaluation format is provided in Fig.\ref{fig:human_evaluation_example}  where the user is asked to rank the performance of each method in the following aspects: 1) color editing of the object, 2) preservation of object identity, 3) preservation of background, and 4) overall editing quality. To prevent potential bias, we use "Method X" as a placeholder for the specific method.

\begin{figure*}[ht]
    \centering
    \includegraphics[width=1.0\textwidth]{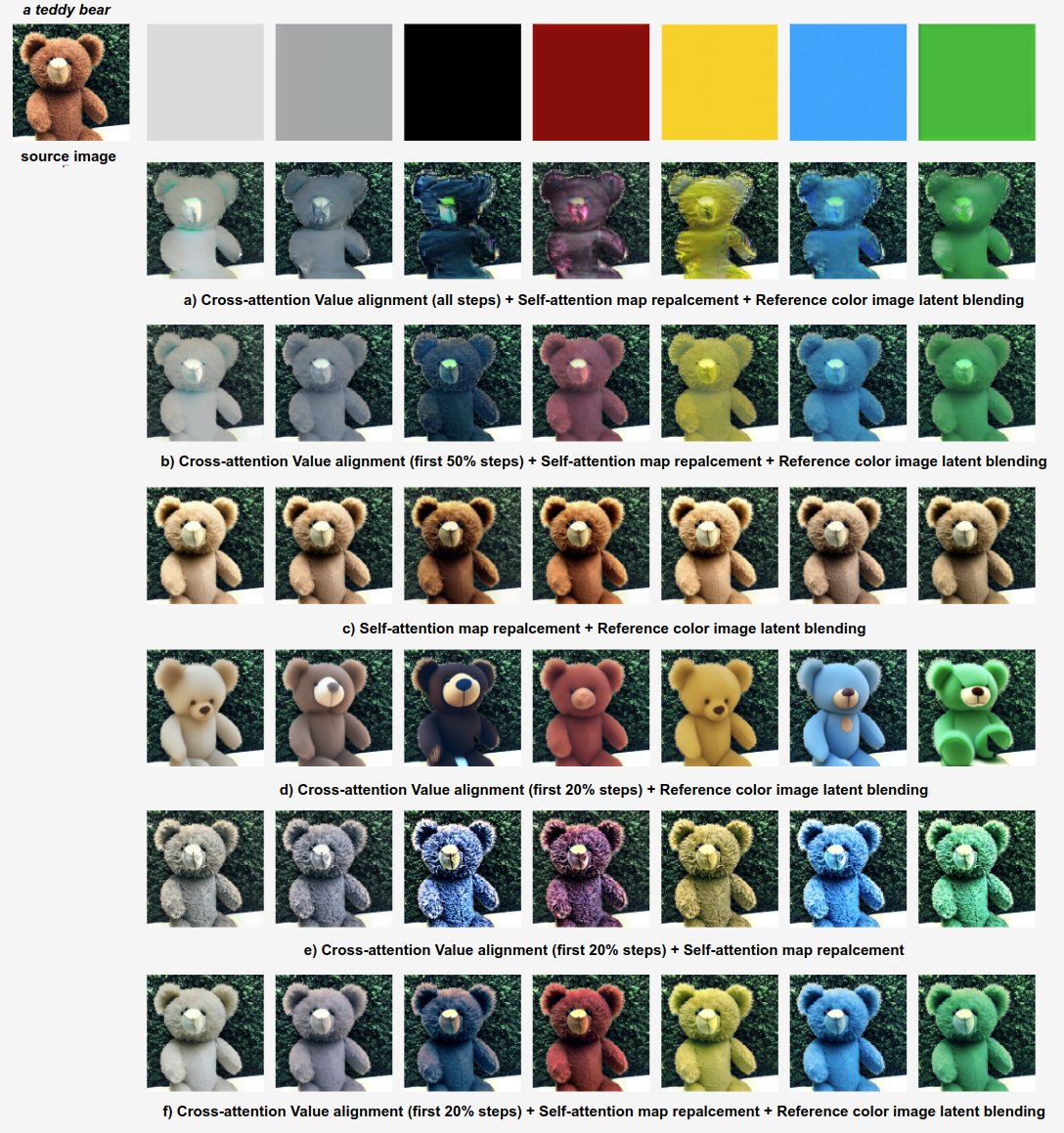}
    \caption{\textbf{Ablation study.}  The original image was generated using the prompt “a teddy bear”, with the object is 'teddy bear'.
  } 
    \label{fig:ablation_study}
\end{figure*}

\textbf{8.7 Ablation Study} 

Fig.\ref{fig:ablation_study} presents the effectiveness of each component in our method. As can be seen in a), b), and f), increasing the number of steps for value matrices alignment in the U-net decoder during the denoising process leads to more substantial changes in the object's identity. Contrasting the results in c) and f) indicates that without value matrices alignment in the cross-attention layer, the object's color remains largely unchanged, varying only in lightness or darkness. Furthermore, as shown in d) and f), the object's structure is altered in the absence of self-attention map replacement. Finally, a comparison of the results in e) and f) demonstrates that the object's texture becomes sharper and its color does not accurately reflect the reference color image when the latent variable of the reference color image $z^{c}_T$ is not blended with the latent variable of the target image $z^{T}_T$.

{
    \small
    \bibliographystyle{ieeenat_fullname}
    \bibliography{main}
}


\end{document}